\documentclass[conference]{IEEEtran}

\usepackage[numbers]{natbib}
\usepackage{multicol}
\usepackage[bookmarks=true]{hyperref}

\usepackage{amsmath}
\usepackage{amsfonts}       % blackboard math symbols
\usepackage{amsthm}         % for proofs
\usepackage{mathtools}
\usepackage{nicefrac}       % compact symbols for 1/2, etc.
\usepackage{times}
\usepackage[super]{nth}
\usepackage{siunitx}
\usepackage{bm}

\usepackage{booktabs}       % professional-quality tables
\usepackage{tabularx}
\usepackage{makecell}

\usepackage{adjustbox}
\usepackage{svg}
\usepackage{microtype}
\usepackage{graphicx}
\usepackage[font=small]{caption}
\usepackage{subcaption}
\usepackage{placeins}

\usepackage[ruled]{algorithm2e} % For algorithms

\SetAlFnt{\small}
\SetAlCapFnt{\small}
\SetAlCapNameFnt{\small}
\SetAlCapHSkip{0pt}

\newcommand{\figref}[1]{Fig.~\ref{#1}}
\newcommand{\secref}[1]{Sec.~\ref{#1}}
\newcommand{\algref}[1]{Alg.~\ref{#1}}
\newcommand{\eqnref}[1]{Eq.~\eqref{#1}}
\newcommand{\tabref}[1]{Tab.~\ref{#1}}
\begin{document}

% paper title
\title{CoCo-InEKF: State Estimation with Learned Contact Covariances in Dynamic, Contact-Rich Scenarios}

\author{
    \IEEEauthorblockN{Michael Baumgartner\IEEEauthorrefmark{1}\IEEEauthorrefmark{2}, David Müller\IEEEauthorrefmark{2}, Agon Serifi\IEEEauthorrefmark{2}, Ruben Grandia\IEEEauthorrefmark{2}, \\
    Espen Knoop\IEEEauthorrefmark{2}, Markus Gross\IEEEauthorrefmark{1}\IEEEauthorrefmark{2}, and Moritz Bächer\IEEEauthorrefmark{2}}
    \IEEEauthorblockA{\IEEEauthorrefmark{1}ETH Zurich, Switzerland, \IEEEauthorrefmark{2}Disney Research, Switzerland}
    %\IEEEauthorblockA{Emails: \{michael.baumgartner, grossm\}@inf.ethz.ch, \{first.last\}@disney.com}
}

\maketitle

\begin{abstract}

Robust state estimation for highly dynamic motion of legged robots remains challenging, especially in dynamic, contact-rich scenarios. Traditional approaches often rely on binary contact states that fail to capture the nuances of partial contact or directional slippage. This paper presents CoCo-InEKF, a differentiable invariant extended Kalman filter that utilizes continuous contact velocity covariances instead of binary contact states. These learned covariances allow the method to dynamically modulate contact confidence, accounting for more nuanced conditions ranging from firm contact to directional slippage or no contact. To predict these covariances for a set of predefined contact candidate points, we employ a lightweight neural network trained end-to-end using a state-error loss. This approach eliminates the need for heuristic ground-truth contact labels. In addition, we propose an automated contact candidate selection procedure and demonstrate that our method is insensitive to their exact placement. Experiments on a bipedal robot demonstrate a superior accuracy-efficiency tradeoff for linear velocity estimation, as well as improved filter consistency compared to baseline methods. This enables the robust execution of challenging motions, including dancing and complex ground interactions --- both in simulation and in the real world.
\end{abstract}

\IEEEpeerreviewmaketitle

\section{Introduction}

Proprioceptive state estimation, i.e., estimating the robot's state without exteroceptive sensors such as cameras or LiDAR, remains a fundamental challenge. These types of estimators provide accurate, high-frequency state information for downstream feedback control during highly dynamic motions or in visually degraded environments. For legged robots, which are typically equipped with an inertial measurement unit (IMU) and actuator encoders that directly measure joint angles, the robot pose and linear velocity estimation requires sensor fusion. 

A common strategy is to estimate the contact state of discrete points on the robot, and assume that points in contact remain stationary in the world frame~\cite{bloesh_state_2012,camurri_pronto_2020,hartley_contact-aided_2020}. The performance of these methods is critically dependent on accurate contact estimation and the validity of the stationarity assumption, motivating specialized methods for contact detection~\cite{hwangbo_probabilistic_2016,camurri_probabilistic_2017} and handling of slipping contacts~\cite{sun_proprioceptive_2025,kim_legged_2021,jenelten_dynamic_2019}. In particular, \citet{lin_legged_2022} proposed augmenting the state-of-the-art invariant extended Kalman filter (InEKF)~\cite{hartley_contact-aided_2020} with learned contact detection. However, this approach requires labeled contact data for training and still treats contact as a binary state, determined independently of the state estimation.

To address these limitations, researchers have explored the use of end-to-end supervised learning for state estimation~\cite{ji_concurrent_2022,yu_state_2024}. Although this approach has enabled the successful deployment of impressive reinforcement learning controllers, these methods underperform when evaluated on state estimation accuracy alone, as we will show in our evaluation.

\begin{figure}[tbp]
  \centering
  \includegraphics[width=\linewidth]{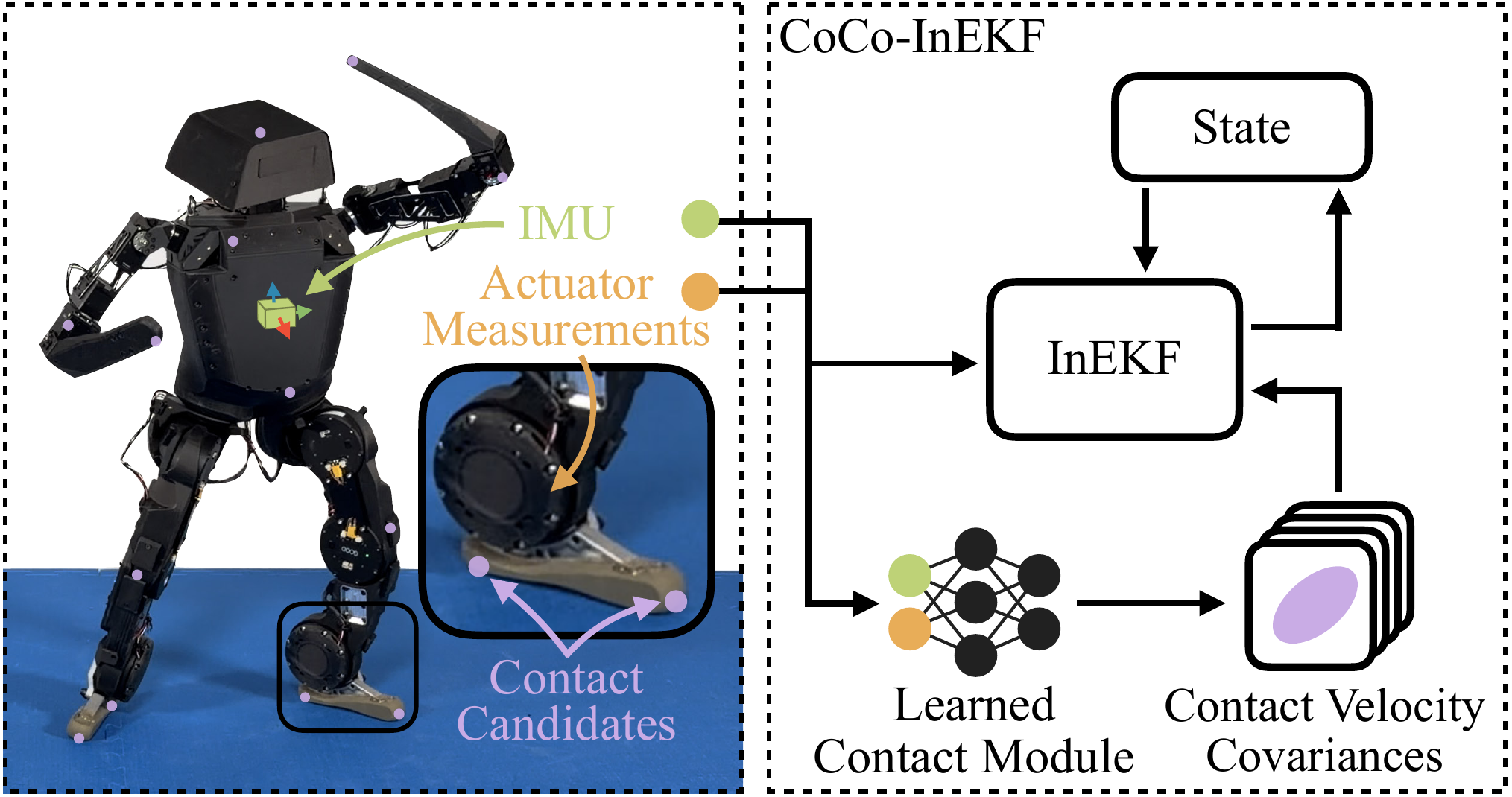} % replace with your file
  \caption{\textbf{CoCo-InEKF.} Given a set of predefined contact candidates and the proprioceptive sensor data from the IMU and actuators of a robot, a learned contact module predicts contact velocity covariances for use in an Invariant EKF.}
  \label{fig:system_overview}
\end{figure}

This paper proposes \textbf{Co}ntact\textbf{Co}variance\textbf{-InEKF}, a novel hybrid approach that combines the strengths of both the end-to-end and InEKF-based methods. We make the standard contact-aided InEKF differentiable by maintaining the state of all contact candidates, regardless of their current contact condition. Motivated by the observation that contact conditions, such as directional slippage, are richer than binary states, we introduce a contact module that predicts contact velocity covariances instead of binary contact states. This allows our model to dynamically modulate contact confidence, which is expressive enough to account for more nuanced states, spanning firm contact, directional slippage, and no contact. However, we lift strict physical enforcement, enabling the model to also build constraints that do not correspond to physical interaction. By applying backpropagation through time (BPTT)~\cite{haarnoja_backprop_2016}, we can then train the neural contact module end-to-end using simple state-error losses, avoiding the need for heuristic ground-truth contact labels required by previous methods. This architecture retains the invariance and observation properties of the InEKF and enables robust state estimation for complex, highly dynamic motions such as dancing and multi-contact ground interactions.

Succinctly, our contributions are: 
\begin{itemize} 
\item CoCo-InEKF, a proprioceptive state estimator combining a differentiable InEKF with a neural contact module that predicts contact velocity covariances.
\item End-to-end training of the proposed architecture via backpropagation through time, avoiding the need for ground-truth contact information.
\item Experimental validation of our method in highly dynamic and contact-rich scenarios, demonstrating improved performance over classical methods, InEKFs with learned contact classification, and end-to-end methods.
\end{itemize}

\section{Related Work}

\subsection{Classical State Estimation}
The pioneering work by \citet{bloesh_state_2012} introduced an EKF that integrates inertial measurements with leg odometry. This approach treats contact points as stationary \textit{landmarks} in global coordinates, using leg forward kinematics to measure their relative locations with respect to the robot's base. This idea has also been integrated into other filtering methods, such as the unscented Kalman filter~\cite{bloesch_state_2013}, factor graphs~\cite{hartley_legged_2018}, and a dual $\beta$-Kalman filter~\cite{zhang_robust_2025}. The InEKF introduced by \citet{hartley_contact-aided_2020} leverages Lie group theory for superior convergence compared to quaternion-based filters. DRIFT~\cite{lin_proprioceptive_2024} provides a modular, modern implementation of this approach.

Independent of the filtering approach, these methods fundamentally rely on an accurately-estimated contact state, and the non-slip assumption. In the absence of direct contact sensors, researchers have developed advanced contact detection methods using force thresholding~\cite{fink_proprioceptive_2020}, probabilistic fusion of kinematics and dynamics~\cite{hwangbo_probabilistic_2016,camurri_probabilistic_2017}, or integration of the planned contact state~\cite{bledt_contact_2018}. Strategies for slip rejection include estimated foot velocity thresholding~\cite{kim_legged_2021}, outlier detection through a threshold on the Mahalanobis distance of the innovation~\cite{bloesch_state_2013}, and estimating slip probability through a hidden Markov model~\cite{jenelten_dynamic_2019}. Recently, \citet{kim_adaptive_2025} adaptively modulated the contact velocity covariance based on an adaptive filtering strategy, an approach similar to ours. While these methods improve robustness, they generally depend on hand-crafted heuristics and user-defined thresholds. In contrast, our approach learns the optimal noise covariance mapping directly from data via a differentiable pipeline, removing the need for expert tuning.

\subsection{Data-Driven State Estimation}

Data-driven approaches to state estimation aim to learn an end-to-end mapping from raw sensor inputs to state trajectories without explicitly encoding a structured filtering algorithm. These methods have seen significant adoption in pedestrian pose estimation, where neural networks process raw IMU data to predict movement~\cite{chen_ionet_2018,herath_ronin_2020}. In the domain of legged robotics, \citet{ji_concurrent_2022} proposed training an estimation network through supervised learning, concurrently with training a reinforcement learning policy. Recently, \citet{yu_state_2024} demonstrated that a transformer architecture can outperform the multilayer perceptron (MLP) used in earlier research. While these black-box methods show promise, they often lack the guarantees and consistency of traditional filters, especially on out-of-distribution data. We utilize the transformer-based approach as a baseline in this work to demonstrate that purely data-driven models do not yet outperform hybrid approaches.

\subsection{Hybrid State Estimation}

Hybrid approaches bridge the gap between physical models and data-driven methods by integrating learning directly into classical filtering frameworks. These strategies are generally categorized by how the learnable components interact with the estimator. One group of methods predicts measurement inputs~\cite{zhang_imu_2021} or input residuals~\cite{brossard_denoising_2020,liu_debiasing_2025} to refine the data before it enters the estimator. Other implementations focus on estimating input noise~\cite{brossard_ai_2020} or relative displacements, which are then fused into the overall state estimate~\cite{liu_tlio_2020,buchanan_learning_2022,cioffi_learned_2023}. These techniques have been widely used in inertial odometry, where networks refine raw IMU data to improve accuracy. Alternative strategies involve predicting state residuals applied to the filter output~\cite{gu_enhancing_2024,lee_legged_2025} or learning the Kalman gain~\cite{liu_-mdf_2023,hohmeyer_inekformer_2025,revach_kalmannet_2022}.

% Differentiablility
The methodology used to train these learnable components represents another critical design choice for hybrid methods. Differentiable formulations enable end-to-end training based solely on state error, eliminating the need for ground-truth labels of intermediate signals~\cite{haarnoja_backprop_2016}. This approach has been successfully applied to various domains, including inertial odometry~\cite{liu_debiasing_2025,brossard_ai_2020,zhang_imu_2021}, manipulation~\cite{lee_multimodal_2020,piga_differentiable_2021,liu_-mdf_2023}, and legged robots~\cite{hohmeyer_inekformer_2025}.

% Legged robotics specific
In the field of legged robotics, many strategies center on the explicit estimation of contact states. \citet{lin_legged_2022} use a network to classify binary contact states based on labeled real-world data. \citet{sun_proprioceptive_2025} learn a slip state and slip velocity estimator, and heuristically modulate the contact velocity covariance based on the predicted slip state. \citet{youm_legged_2025} learn to predict the robot's base linear velocity in addition to contact states. Although these learning-based contact estimators improve performance over their model-based counterparts, they often still rely on thresholds and heuristics to obtain the ground-truth data necessary for supervised learning or to derive a binary contact prediction from a continuous network output. Our method addresses this by learning contact velocity covariances end-to-end via a differentiable formulation.

\section{Method}

The aim of this work is to enable robust state estimation across a variety of challenging contact-rich scenarios given a robot morphology, including the IMU location, and a user-specified set of contact candidate points on the robot model. As illustrated in~\figref{fig:system_overview}, at each time step, a learned module predicts a covariance for each contact candidate point, which encodes the (directional) confidence that the point is stationary in the world frame. These predictions are then fed into a standard InEKF formulation that fuses IMU measurements and leg odometry.

CoCo-InEKF is based on the contact-aided InEKF~\cite{hartley_contact-aided_2020}, which estimates the following state, with the world frame denoted by $W$, the IMU (body) frame by $B$, and each contact frame by $C_i$
\begin{equation*}
\mathbf{x} := \left( {}^{W}\mathbf{R}_{B}, {}^{W}\mathbf{v}_B, {}^{W}\mathbf{p}_{B}, {}^{W}\mathbf{p}_{C_1}, \dots , {}^{W}\mathbf{p}_{C_N}, {}^{B}\mathbf{b}_{\bm{\omega}}, {}^{B}\mathbf{b}_{\mathbf{a}} \right).
\end{equation*}
The state consists of base orientation ${}^{W}\mathbf{R}_{B}$, base linear velocity ${}^{W}\mathbf{v}_B$, base position ${}^{W}\mathbf{p}_{B}$, $N$ contact positions ${}^{W}\mathbf{p}_{C_i}$, and IMU gyro and acceleration biases ${}^{B}\mathbf{b}_{\bm{\omega}}$ and ${}^{B}\mathbf{b}_{\mathbf{a}}$. In the standard contact-aided approach, the contact positions are dynamically added to the state when contact is detected, and removed when contact is released. However, in our approach, we permanently maintain the positions of all contact candidates as part of the state, resulting in a differentiable\footnote{Without the discrete addition and removal of contact points, the InEKF consists of a constant computation graph, where all operations (matrix multiplication, inversion, and group operations) are differentiable.} InEKF.

This key difference in modeling is further highlighted by examining the process model for the contact positions,
\begin{equation}
    {}^{W}\dot{\mathbf{p}}_{C_i} = -{}^{W}\mathbf{R}_{B} {}^{B}\mathbf{w}_{C_i},
\end{equation}
where ${}^{B}\mathbf{w}_{C_i} \sim \mathcal{N}(\bm{0}_{3 \times 1}, {}^{B}\bm{\Sigma}_{C_i})$ is a Gaussian noise term\footnote{Compared to the formulation in~\cite{hartley_contact-aided_2020}, we formulate the noise term directly in the IMU frame instead of the contact frame.}. This zero-mean velocity model for the contact positions motivates the original interpretation of static landmarks. However, in our method, where a learned module predicts covariances ${}^{B}\bm{\Sigma}_{C_i}$, the model's confidence is continuously adjusted. The states ${}^{W}\mathbf{p}_{C_i}$ should therefore rather be interpreted as the continuous positions of the contact candidates in global coordinates, independent of their contact condition.

\subsection{Contact-Aided InEKF}

Apart from the deviation discussed in the previous section, we follow the standard contact-aided InEKF formulation as summarized in high-level pseudocode in \algref{alg:inekf_short}, including the contact covariances predicted by our learned module. Because our formulation changes are limited to keeping all contacts in the state at all times, we retain the desirable filter invariance and observability properties: given at least one contact point, the state is observable up to global translation and yaw rotation~\cite{hartley_contact-aided_2020}.
We omit the derivation of the linearization and discretization of the filter's process and measurement models, providing only their continuous-time, non-linear equations in the subsequent sections. For an in-depth discussion, see \citet{hartley_contact-aided_2020}.

\paragraph*{Prediction}
The continuous, non-linear process model for the state $\mathbf{x}$ is given by
\begin{align}
{}^{W}\dot{\mathbf{R}}_{B} &= {}^{W}\mathbf{R}_{B} \left({}^{B}\bm{\omega} - {}^{B}\mathbf{b}_{\bm{\omega}} - {}^{B}\mathbf{w}_{\bm{\omega}}\right)_\times, \label{eq:dynamics:R} \\
{}^{W}\dot{\mathbf{v}}_B &=  {}^{W}\mathbf{R}_{B}  \left({}^B\mathbf{a} - {}^{B}\mathbf{b}_\mathbf{a} - {}^{B}\mathbf{w}_{\mathbf{a}} \right) + {}^{W}\mathbf{g}, \label{eq:dynamics:v}  \\
{}^{W}\dot{\mathbf{p}}_{B} &= {}^{W}\mathbf{v}_B, \label{eq:dynamics:p} \\
{}^{W}\dot{\mathbf{p}}_{C_i} &= -{}^{W}\mathbf{R}_{B} {}^{B}\mathbf{w}_{C_i}, \label{eq:dynamics:p_c} \\
{}^{B}\dot{\mathbf{b}}_{\bm{\omega}} &= {}^{B}\mathbf{w}_{\mathbf{b}_{\bm{\omega}}}, \label{eq:dynamics:b_omega} \\ 
{}^{B}\dot{\mathbf{b}}_{\mathbf{a}} &= {}^{B}\mathbf{w}_{\mathbf{b}_\mathbf{a}}, \label{eq:dynamics:b_a}
\end{align}
where Eqs.~\eqref{eq:dynamics:R}--\eqref{eq:dynamics:p} model the IMU state forward integration based on gyro measurements ${}^{B}\bm{\omega}$, accelerometer measurements ${}^B\mathbf{a}$, and the gravity vector ${}^{W}\mathbf{g}$. Eq.~\eqref{eq:dynamics:p_c} models the zero-mean contact candidate velocity, and Eqs.~\eqref{eq:dynamics:b_omega}~and~\eqref{eq:dynamics:b_a} model random walks for the bias terms. The additive noise terms ${}^{B}\mathbf{w}_{\bm{\omega}}$, ${}^{B}\mathbf{w}_{\mathbf{a}}$, $ {}^{B}\mathbf{w}_{\mathbf{b}_{\bm{\omega}}}$, and $ {}^{B}\mathbf{w}_{\mathbf{b}_\mathbf{a}}$ are zero-mean Gaussians for gyro measurements, linear acceleration measurements, and bias drift.

\paragraph*{Correction} The leg kinematics are used to form a measurement model. Given joint position measurements $\mathbf{q}$, the forward kinematics $\mathbf{h}_{C_i}(\mathbf{q})$ returns the position of frame $C_i$ relative to the body frame $B$, expressed in $B$. The same quantity can be derived from the estimated state, leading to the measurement equation
\begin{equation}
    \mathbf{h}_{C_i}(\mathbf{q}) = {}^{W}\mathbf{R}_{B}^\top \left({}^{W}\mathbf{p}_{C_i} - {}^{W}\mathbf{p}_{B} \right) + {}^{B}\mathbf{J}_{C_i}(\mathbf{q}) \mathbf{w}_{\mathbf{q}},
    \label{eq:kin_meas}
\end{equation}
where ${}^{B}\mathbf{J}_{C_i}(\mathbf{q})$ is the Jacobian of the forward kinematics, and $\mathbf{w}_{\mathbf{q}}$ is a zero-mean Gaussian noise for the joint positions.

\begin{algorithm}[tb]
\caption{CoCo-InEKF. The $\oplus$ notation signifies an addition on the state manifold.}
\label{alg:inekf_short}
\LinesNumbered
\SetAlgoLined % Adds the vertical lines for loops
\DontPrintSemicolon % Hides the semicolons at the end of lines in the PDF

\textbf{Initialize:} $\mathbf{x}, \mathbf{P}$\;

\For{each time step $t$}{
    \textbf{Input:} IMU and actuator measurements\;
    
    \tcp*[l]{1. Prediction}

    $^B\bm{\Sigma}_{C_i} \leftarrow$ \text{ContactNet} \;
    $\bm{\Phi}, \mathbf{Q} \leftarrow$ Linearization \& discretization of Eqs.~\eqref{eq:dynamics:R}--\eqref{eq:dynamics:b_a}\;
    $\mathbf{x}^- \leftarrow$ Integrate Eqs.~\eqref{eq:dynamics:R}--\eqref{eq:dynamics:b_a} \;
    $\mathbf{P}^- \leftarrow \bm{\Phi} \mathbf{P} \bm{\Phi}^\top + \mathbf{Q}$\;   
    
    \tcp*[l]{2. Correction}
    $\mathbf{z}, \mathbf{H}, \mathbf{N} \leftarrow$ Linearized measurement \eqnref{eq:kin_meas}\;
    $\mathbf{K} \leftarrow \mathbf{P}^- \mathbf{H}^\top (\mathbf{H} \mathbf{P}^- \mathbf{H}^\top + \mathbf{N})^{-1}$ \;
    $\mathbf{x} \leftarrow \mathbf{K} \mathbf{z} \oplus \mathbf{x}^-$\;
    $\mathbf{P} \leftarrow (\mathbf{I} - \mathbf{K} \mathbf{H}) \mathbf{P}^-$\;
}
\end{algorithm}

\subsection{Learning Contact Covariances}

Our learned contact module predicts per-contact-point velocity covariances in the body frame ${}^{B}\bm{\Sigma}_{C_i}$. We ensure that the output of the learned contact module is a valid, symmetric, positive-semidefinite matrix by predicting the $6$ elements of a lower-triangular $3\times3$ matrix $\mathbf{L}$. The covariance is then given by ${}^{B}\bm{\Sigma}_{C_i} = \mathbf{L}\mathbf{L}^\top$. The input to this module is a history of length $H$ of proprioceptive sensor inputs, 
\begin{equation*}
    \mathbf{o} := \left({}^{B}\bm{\omega}, {}^{B}\mathbf{a}, \mathbf{q}, \dot{\mathbf{q}}, \bm{\tau}, {}^{B}\mathbf{p}_{B\rightarrow C_i}, {}^{B}\mathbf{v}_{B\rightarrow C_i}\right),
\end{equation*}
where $\bm{\tau}$ are actuator torque measurements, ${}^{B}\mathbf{p}_{B\rightarrow C_i} = \mathbf{h}_{C_i}(\mathbf{q})$ are relative contact positions, and ${}^{B}\mathbf{v}_{B\rightarrow C_i}$ are their relative velocities. To maintain a clean separation between the filter dynamics and learned predictions, and to avoid complex feedback loops, we intentionally exclude internal InEKF state estimates from the contact module's input features. The history of $H$ inputs is denoted by $\mathbf{O}$. As in~\cite{lin_legged_2022}, these inputs are normalized along the time dimension.

Due to computational constraints, we use the more efficient MLP implementation introduced in \cite{lin_legged_2022}, as the convolutional neural network (CNN)-based approach is unable to run in real time on our onboard system. We evaluate the impact of this architectural change in~\secref{sec:arch_choices}.

\begin{figure}[tbp]
  \centering
  \includegraphics[width=\linewidth]{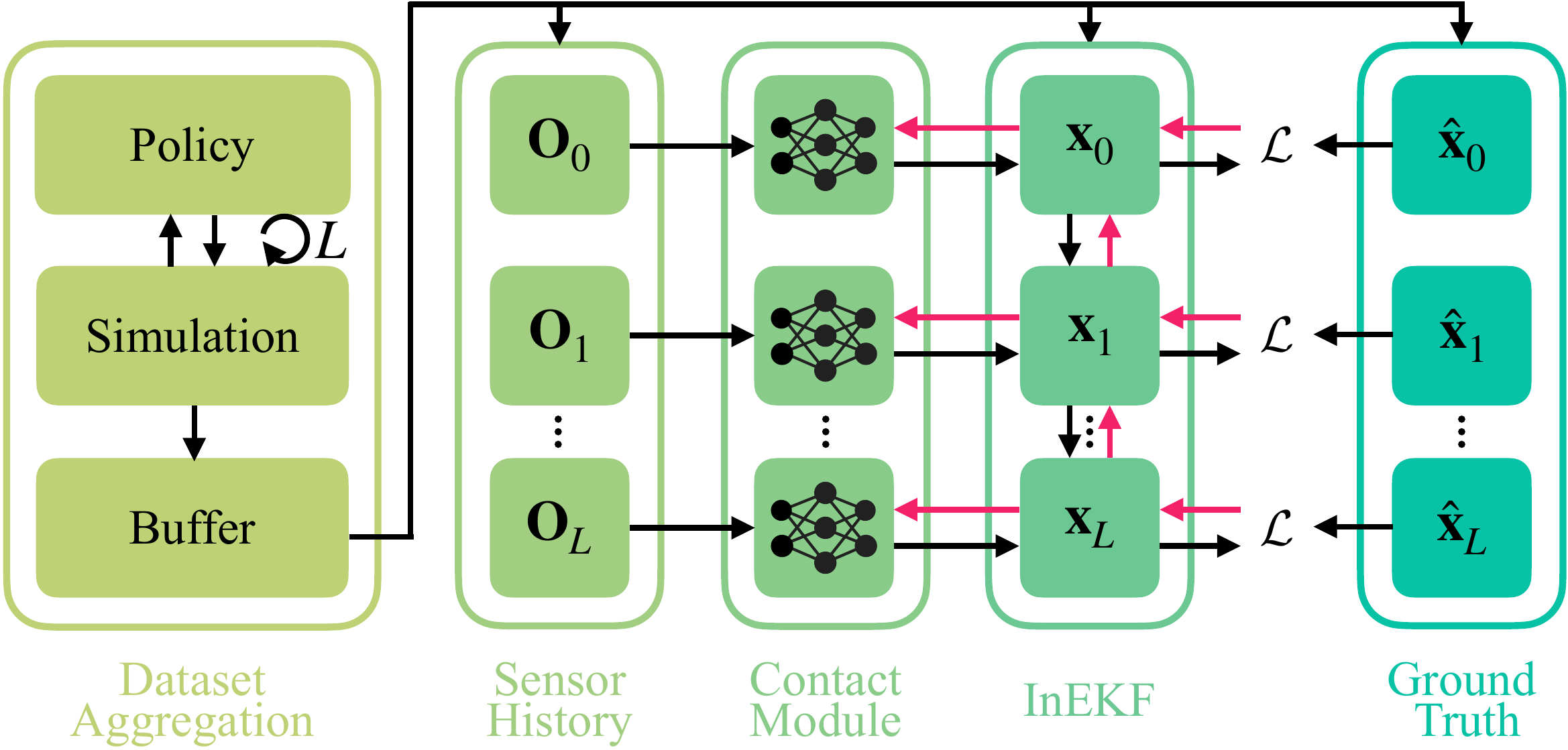}
  \caption{\textbf{Training setup}. Per learning iteration, we collect a training dataset of ground-truth states and sensor measurements by forward-simulating a pretrained policy in $E$ environments. The contact module predicts contact velocity covariances based on the sensor history (IMU + actuator data). These covariances are passed to the differentiable InEKF, rolling out the state estimate for $L$ time steps. We compute a loss on the error between the estimated and ground-truth states, and backpropagate gradients (red arrows) from all time steps through the InEKF to the contact module parameters.
  }
  \label{fig:training_setup}
\end{figure}

Our training setup uses BPTT, as illustrated in \figref{fig:training_setup}.
Given an existing control policy, we record the sensor input history and ground-truth states as we roll out the policy in simulation. 
This is done in parallel across $E$ environments, where we randomize friction, terrain shapes, and disturbance forces.
At environment initialization, the InEKF is set to the ground-truth state and an initial covariance. Afterward, the InEKF maintains its internal state and covariance across consecutive rollouts up to a maximum episode length $T$, allowing it to drift away from the ground-truth state. 
During a rollout, the proprioceptive and ground-truth state data are stored in a buffer of length $L$ and are then used to run CoCo-InEKF and compute losses by comparing the estimated and ground-truth states.
Finally, we average the loss across time and environments, backpropagate through our differentiable InEKF to each contact covariance prediction, and update the contact module parameters using the Adam optimizer for a learning iteration.
We experimented with re-initializing the InEKF state to the ground-truth state at the start of each rollout in a force-teacher fashion, but observed a degradation in the filter's learning performance.

As the loss function, we use an L2 loss on the body velocity in the body frame,
\begin{equation}
    \mathcal{L}\left(\mathbf{x}, \hat{\mathbf{x}}\right) = \left\| {}^{W}\mathbf{R}_{B}^\top{}^{W}\mathbf{v}_{B} - {}^{W}\hat{\mathbf{R}}_{B}^\top{}^{W}\hat{\mathbf{v}}_{B} \right\|^2_2,
    \label{eq:loss}
\end{equation}
where the hat symbol ($\hat{\cdot}$) indicates ground-truth values from the simulator. We optimize for the velocity in the body frame, as this is the frame used by the downstream control policies. Moreover, the body frame velocities are an observable quantity of the InEKF, while the velocity in the world frame is only partially observable due to the unobservable global yaw of the robot. This quantity is also not affected by the InEKF's drifting behavior during an episode. Adding additional loss terms based on the position and orientation states is straightforward, however the InEKF's global drift over the episode must be accounted for in the loss formulations.

\subsection{Automated Contact Candidate Selection}

To facilitate the application of CoCo-InEKF to new robots, we present an automated method for generating a set of contact candidates. The method's performance is on par with that of hand-picked contact point sets, as shown in \secref{sec:ablation_contact_point_configuration}.

Given a target number of $N$ contact candidates, and a set of robot rigid bodies on which contact points should be placed, we sample $10\cdot N$ points per rigid body on their respective mesh surfaces, concatenating them. Starting from a random seed and placing the robot in a nominal pose, we then use farthest point sampling to pick $N$ points. Intuitively, this will place points at extremities such as the toes or hands, as well as distributing them across the robot.

\section{Experimental Setup}

We test CoCo-InEKF against several baselines, including state-of-the-art state estimation approaches. For a fair comparison, we adapt the baseline methods to accommodate the limited computing power of our robot hardware, evaluating multiple baseline variations for completeness. Models are trained and evaluated on two scenarios consisting of different motion types: dynamic \textit{dancing motions} (foot contact only) and \textit{ground motions} (full-body ground contact). We evaluate design choices through a set of ablation studies, investigate the impact of framework changes on the statistical consistency of the filter, and demonstrate that our method enables dynamic motion on a physical robot.

\subsection{Implementation Details}

We evaluate models on Lima, a custom bipedal robot (\SI{0.84}{\meter}, \SI{16.2}{\kilo\gram}, 20 DoF) with an onboard computer (Intel i7, 4-Core, \SI{1.7}{\giga\hertz}) running a \SI{600}{\hertz} control loop. For dancing motions, we place contact points at the heels and toes for a total of $N=4$. For ground motions, we place $N=10$ points across the robot. 

Models are trained on a single Nvidia RTX 4090 GPU for 100k iterations, or a maximum of 5 days, with $E=1280$ parallel environments. Unless otherwise noted, we use a history of $H=20$ and a training buffer length of $L=128$.  All simulations are performed at \SI{600}{\hertz}, matching the robot's low-level loop rate. 

For computational timing benchmarks, the models are executed single-threaded on the robot’s onboard computer as part of the real-time control loop.

\subsubsection{Baseline Methods}

We evaluate CoCo-InEKF against a set of approaches, as listed in \tabref{tab:baselines}. The InEKF method with ground-truth contacts cannot be implemented on a real robot, but presents a best-case scenario for an InEKF with binary contacts.

\begin{table}
\centering
    \caption{Baseline methods.}
    \label{tab:baselines}
    \footnotesize
    \begin{tabular}{p{0.35\linewidth} p{0.5\linewidth}}
    \toprule
    \textit{InEKF, GT contacts} & InEKF, with ground-truth contact using xy-velocity (\SI{\leq0.25}{\meter/\second}) and height (\SI{\leq0.01}{\meter}). \\
    \textit{InEKF, heuristic contacts} & InEKF, with contact heuristic using estimated xy-velocity (\SI{\leq0.25}{\meter/\second}) and height (\SI{\leq0.01}{\meter}). \\
    \textit{Hybrid Baseline} &  Method from \cite{lin_legged_2022}. \\
    \textit{Hybrid Baseline+} & As Hybrid Baseline, but with reduced model size and added slip classification. \\
    \textit{SET} & Unstructured end-to-end transformer-based method \cite{yu_state_2024}. \\
    \bottomrule
    \end{tabular}
\end{table}

For the Hybrid Baseline+ method, we adapt the Hybrid Baseline method from \citet{lin_legged_2022} to run on our available compute by: (1) removing CNN layers and flattening input data; (2) reducing the MLP hidden layer sizes to 128 and 64 units; (3) predicting the contact state of each contact point independently; and (4) adding a `slip' prediction class. Similar to \citet{kim_adaptive_2025}, we then increase the contact velocity covariance by a factor of $10\times$ when slip is predicted. As we show in \secref{sec:arch_choices}, these changes yield a significant speedup while maintaining performance.

For the state estimation transformer (SET) method from \citet{yu_state_2024}, we adapt the inputs to be consistent with our model and express linear velocities in the body frame so that our loss formulation can be used. We integrate velocity predictions using forward Euler to obtain world position estimates, as these are not directly estimated by the method, and use complementary IMU filtering \cite{valenti_keeping_2015} for a more robust orientation estimate. To evaluate performance trade-offs of this approach, we compare two model sizes as listed in \tabref{tab:SET_small_large}.

\begin{table}[tb]
    \centering{}
    \caption{Configuration of small and large SET models.}
    \label{tab:SET_small_large}
    \footnotesize
    \begin{tabular}{l r r}
    \toprule
     & \textbf{Small} & \textbf{Large} \\
     \midrule
     Self-attention blocks & 6 & 6 \\
     Heads per block & 4 & 8 \\
     Linear token embedding dimensions & 128 & 256 \\
     MLP hidden dimensions & 256 & 1024 \\
     \bottomrule
     \end{tabular}
     
\end{table}

\subsection{Datasets}

For the dancing scenario, we use a VMP policy \cite{serifi_vmp_2024} that tracks arbitrary kinematic reference motions. The observations follow \cite{serifi_vmp_2024} with an asymmetric actor-critic training setup where the privileged observations for the critic are noise-free. For the reference motions, we use a subset of the Reallusion dataset \cite{reallusion}, with 81 sequences of \SI{5.6}{}--\SI{36.1}{\second} duration, retargeted onto the Lima robot. For training, we concatenate randomly-drawn sequences from this dataset and track them in simulation up to an episode length $T=$ \SI{100}{\second}. The dancing test dataset comprises $100$ such sequences, each \SI{20}{\second} in duration, drawn from the same motion pool but concatenated differently.

For the ground motions, we use a stylized falling policy \cite{strauch_robot_2025} to track a sequence of goal end poses. For each episode, the robot starts from a standing pose, the goal pose changes every \SI{2}{\second}, and the environment is reset after an episode length $T=$ \SI{6}{\second} during training. This produces contact-rich motions with challenging full-body ground contacts. The ground motion test dataset again comprises 100 such sequences, each \SI{20}{\second} in duration, produced by different concatenations of the same goal poses to include more ground interactions.

During training, we apply domain randomization of friction coefficients and disturbance forces. In the dancing scenario, we also vary the terrain. For the test datasets, we apply the same randomization of friction coefficients, but omit the terrain variation and only include periodic disturbances for the ground motions to induce slippage.

\subsection{Metrics}

Offline metrics are based on \cite{zhang_tutorial_2018}. We report absolute trajectory error (ATE), computed over the trajectory after aligning the initial state, as well as the root mean square error (RMSE), mean absolute error (MAE), median absolute error (MED), and standard deviation (STD). Linear velocity errors are computed in the robot IMU body frame, to prevent cross-coupling to orientation errors. Error units are \SI{}{\meter/\second}, \SI{}{\meter}, \SI{}{\radian} for velocity, position, and rotation, respectively.

For the statistical filter analysis, we evaluate the normalized estimation error squared (NEES) \cite{BarShalom2001} for the combined core filter states comprised of position, orientation, and velocity, as well as for the states in isolation.

\section{Results}

\begin{figure}[tbp]
  \centering
  \includegraphics[width=\linewidth]{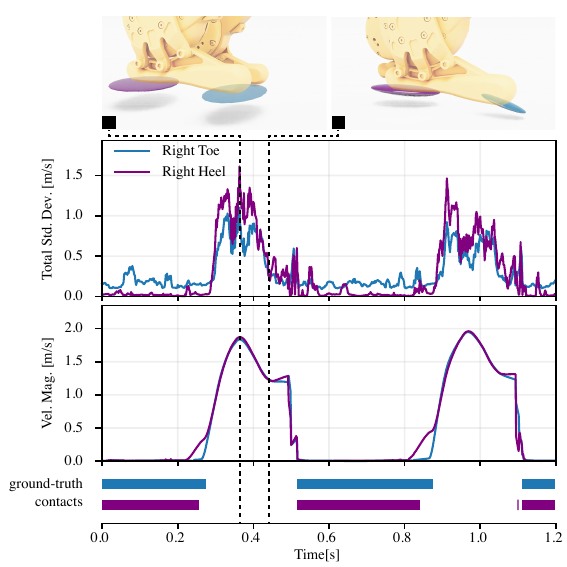}
  \caption{\textbf{Contact Covariance.} Visualization of the contact covariance total standard deviation $\sqrt{\text{tr}(^{B}\bm{\Sigma}_{{}C_i{}})}$ and contact point velocity magnitude during a forward-walking gait of our robot, along with ground-truth contacts.
  }
  \label{fig:contact_covariance_gait_viz}
\end{figure}

\subsection{Simulation Experiments}

\begin{table*}[tb]
\caption{ATE comparison on simulated dancing motions.}
\label{tab:ate_total_comparison_dancing}
\centering
\footnotesize
\begin{tabular}{lrrrrrrrrrrrr}
\toprule
& \multicolumn{4}{c}{\textbf{Linear Velocity ATE}} & \multicolumn{4}{c}{\textbf{Position ATE}} & \multicolumn{4}{c}{\textbf{Orientation ATE}}\\
\cmidrule(lr){2-\numexpr5\relax} \cmidrule(lr){6-\numexpr9\relax} \cmidrule(lr){10-\numexpr13\relax}
\textbf{Model}  & RMSE & MAE & MED & STD & RMSE & MAE & MED & STD & RMSE & MAE & MED & STD \\
\midrule
InEKF, GT contacts & 0.176 & 0.070 & 0.037 & 0.162 & 0.422 & 0.119 & 0.042 & 0.405 & 0.033 & 0.022 & 0.013 & 0.025 \\
InEKF, heuristic contacts & 2.675 & 1.429 & 0.436 & 2.262 & 17.423 & 8.260 & 0.866 & 15.340 & 0.041 & 0.031 & 0.023 & 0.027 \\
Hybrid Baseline & 0.123 & 0.060 & 0.034 & 0.107 & 0.163 & 0.078 & 0.048 & 0.143 & 0.031 & 0.019 & \textbf{0.011} & 0.025\\
Hybrid Baseline+ & 0.121 & 0.061 & 0.036 & 0.105 & \textbf{0.111} & \textbf{0.065} & \textbf{0.040} & \textbf{0.090} & \textbf{0.027} & \textbf{0.019} & 0.012 & \textbf{0.020}\\
CoCo-InEKF (ours) & \textbf{0.046} & \textbf{0.028} & \textbf{0.018} & \textbf{0.037} & 0.124 & 0.079 & 0.052 & 0.096 & 0.031 & 0.020 & 0.013 & 0.024\\
SET, small & 0.279 & 0.195 & 0.138 & 0.199 & 0.487 & 0.379 & 0.313 & 0.305 & 0.072 & 0.059 & 0.052 & 0.042 \\
SET, large & 0.286 & 0.203 & 0.143 & 0.202 & 0.395 & 0.290 & 0.205 & 0.269 & 0.072 & 0.059 & 0.052 & 0.042 \\
\bottomrule
\end{tabular}
\end{table*}

\begin{table*}[tb]
\caption{ATE comparison on simulated ground motions.}
\label{tab:ate_total_comparison_ground_motion}
\centering
\footnotesize
\begin{tabular}{lrrrrrrrrrrrr}
\toprule
 & \multicolumn{4}{c}{\textbf{Linear Velocity ATE}} & \multicolumn{4}{c}{\textbf{Position ATE}} & \multicolumn{4}{c}{\textbf{Orientation ATE}} \\
\cmidrule(lr){2-\numexpr5\relax} \cmidrule(lr){6-\numexpr9\relax} \cmidrule(lr){10-\numexpr13\relax}
\textbf{Model} & RMSE & MAE & MED & STD & RMSE & MAE & MED & STD & RMSE & MAE & MED & STD \\
 \midrule
InEKF, GT contacts & 0.418 & 0.296 & 0.203 & 0.295 & 1.573 & 0.971 & 0.512 & 1.238 & 0.078 & 0.048 & 0.028 & 0.061 \\
InEKF, heuristic contacts & 4.448 & 3.110 & 2.117 & 3.179 & 32.695 & 19.738 & 10.248 & 26.065 & \textbf{0.049} & 0.039 & 0.032 & \textbf{0.029} \\
Hybrid Baseline & 0.363 & 0.266 & 0.190 & 0.247 & 1.455 & 0.899 & 0.414 & 1.144 & 0.063 & \textbf{0.037} & \textbf{0.020} & 0.051 \\
Hybrid Baseline+& 0.428 & 0.316 & 0.232 & 0.289 & 1.982 & 1.238 & 0.613 & 1.547 & 0.074 & 0.041 & 0.021 & 0.061 \\
CoCo-InEKF (ours) & 0.099 & 0.069 & 0.042 & \textbf{0.071} & 0.342 & 0.271 & 0.239 & 0.210 & 0.069 & 0.056 & 0.049 & 0.040 \\
SET, small & 0.107 & 0.069 & 0.034 & 0.082 & \textbf{0.126} & \textbf{0.100} & \textbf{0.087} & \textbf{0.076} & 0.058 & 0.041 & 0.031 & 0.041 \\
SET, large & \textbf{0.096} & \textbf{0.063} & \textbf{0.032} & 0.072 & 0.159 & 0.132 & 0.124 & 0.088 & 0.058 & 0.041 & 0.031 & 0.041 \\
\bottomrule
\end{tabular}
\end{table*}

\subsubsection{Contact Covariances} 
To provide insights into the velocity covariance representation, we analyze a forward-walking gait cycle, see \figref{fig:contact_covariance_gait_viz}. We plot the total covariance standard deviation and compare it to the ground-truth binary contact state (computed as in \tabref{tab:baselines}), for the right heel and toe contact points. We also visualize the covariance ellipsoids for two instances in time. 

Although physical interpretability is not enforced by our framework, we observe an alignment of the ground-truth contacts with features in the covariance. However, the covariance also includes additional nuances that the InEKF can leverage. For example, the total standard deviations of the covariance ellipsoids are observed to agree with the instantaneous velocities of the contact points. Moreover, inspecting two snapshots of the forward gait, the directionality of the covariances can be seen to generally be in agreement with the contact candidates' displacement, constraining the tangential directions via low covariance magnitudes.

\subsubsection{Dancing Motions}

The metrics for CoCo-InEKF and the baseline methods, when trained and evaluated on the simulated dancing motions, are summarized in \tabref{tab:ate_total_comparison_dancing}. Our proposed CoCo-InEKF model achieves the lowest linear velocity errors, while the InEKF approach with heuristic contact detection exhibits very high errors, indicating estimator divergence. While our method is slightly outperformed in terms of position and orientation ATE, we hypothesize that adding additional losses on these states could improve our method's performance on those metrics. Because of the differentiable training setup, adding such loss terms to \eqnref{eq:loss} is trivial.

\subsubsection{Ground Motions}

The linear velocity ATE in root frame as well as the position and orientation ATE in world frame are summarized in \tabref{tab:ate_total_comparison_ground_motion} for the simulated ground motions. Our CoCo-InEKF matches the performance of the much larger SET models for 10 contact points, and shows a significant reduction in all error metrics compared to our other baselines. However, as we show in \secref{sec:num_cp_scaling_ablation}, compared to SET, our novel approach is significantly faster, and able to better scale to additional contact points on the robot body.

Note that 10 contact candidates results in a comparatively sparse contact set for full-body motions, as can be seen from \figref{fig:contact_configurations_full_body}. We hypothesize that as our method reasons about contacts in terms of velocity covariances rather than binary contact flags, the contact points are not required to be exactly in contact. As soon as a contact candidate is stationary along a certain dimension, this information can be used to improve the state estimate.

\subsection{Ablation Studies}
\label{sec:ablations}

To analyze the contribution of our design choices to the overall performance of CoCo-InEKF, we conduct a series of ablation studies focusing on network architecture and the specific hyperparameters used during training.

\subsubsection{Architecture and Inputs}
\label{sec:arch_choices}

We first evaluate the performance and inference time of different network architectures. We compare the Hybrid Baseline (CNN architecture), Hybrid Baseline+, SET baselines, and our method. We also evaluate the effect of each of the changes that were made between Hybrid Baseline and Hybrid Baseline+: replacing the CNN with an MLP architecture; reducing MLP hidden layer sizes and predicting each contact independently; adding `slip' classification to the outputs. The results are summarized in \tabref{tab:ablation_comparison_model_size}. It can be seen that Hybrid Baseline+ achieves better performance than Hybrid Baseline, and a $5\times$ speedup. The SET baselines are computationally most expensive, with the highest RMSE. CoCo-InEKF clearly outperforms the other methods.

We also investigate the effect of history size, comparing $H=20$ to $H=150$. Results are shown in \tabref{tab:ablation_comparison_history_size}. For all methods, $H=150$ causes performance degradation both in terms of inference time as well as estimation accuracy.

\begin{table}[tb]
\caption{Model architecture ablation w.r.t. linear velocity ATE on dancing motions for our model, baseline models, and intermediate models between Hybrid Baseline and Hybrid Baseline+, showing the effects of individual changes. We also report the number of parameters and the inference time for the neural network (NN), together with the full state estimator (SE). For SET, a single value is reported, as the SE consists solely of the NN.}
\label{tab:ablation_comparison_model_size}
\centering
\footnotesize
\begin{tabular}{lrrrr}
\toprule
\textbf{Model}
 & RMSE & \# Params. & NN / SE [ms] \\
\midrule
Hybrid Baseline & 0.123 & 2'473'744 & 1.81 / 2.10  \\
Hybrid Baseline, MLP & 0.135 & 239'824 & 0.15 / 0.44 \\
Hybrid Baseline+ w/o slip & 0.122 & 239'304 & 0.16 / 0.45  \\
Hybrid Baseline+ & 0.121 & 239'564 & 0.16 / 0.45 \\
CoCo-InEKF (ours) & \textbf{0.046} & 240'344 & 0.14 / 0.42   \\
SET, small & 0.279 & 810'755 & 2.09 \\
SET, large & 0.286 & 4'770'307 & 3.02  \\
\bottomrule
\end{tabular}
\end{table}

\begin{table}[tb]
\caption{History size ablation, $H=20$ vs. $H=150$. We also report number of parameters and the inference time for the neural network (NN) and the full state estimator (SE), respectively.}
\label{tab:ablation_comparison_history_size}
\centering
\footnotesize
\begin{tabular}{lrrrrr}
\toprule
\textbf{Model}
 & $H$ & RMSE & \# Params. & NN / SE [ms] \\
\midrule
Hybrid Baseline & 20 & 0.123 & 2'473'744 & 1.81 / 2.10 \\
Hybrid Baseline & 150 & 0.150 & 10'862'352 & 6.43 / 6.79 \\
Hybrid Baseline+ & 20 & 0.121 & 239'564 & 0.16 / 0.45 \\
Hybrid Baseline+ & 150 & 0.124 & 1'737'164 & 0.62 / 0.98 \\
CoCo-InEKF & 20 & \textbf{0.046} & 240'344 & 0.14 / 0.42 \\
CoCo-InEKF & 150 & 0.052 & 1'737'944 & 0.61 / 0.97 \\
\bottomrule
\end{tabular}
\end{table}

\subsubsection{Training Configurations}

We study the effect of changing the BPTT horizon $L$, for lengths 64, 128, and 256. Results are summarized in \tabref{tab:ablation_comparison_BPTT}. Significant performance deterioration is seen for the shorter buffer size, presumably because the model cannot predict longer-horizon effects. Performance is also slightly worse for the longer buffer size, which could be explained by vanishing or exploding gradients, or due to higher training cost leading to fewer training iterations.

\begin{table}[tb]
\caption{BPTT unroll size ablation. We report linear velocity ATE on synthetic dancing data, as well as the number of training iterations (limited by the 5-day training time).}
\label{tab:ablation_comparison_BPTT}
\centering
\footnotesize
\begin{tabular}{lrrrrr}
\toprule
 & RMSE & MAE & MED & STD & \# Iters. \\
\midrule
$L=64$ & 0.066 & 0.043 & 0.027 & 0.051 & 89'600\\
$L=128$ (ours) & \textbf{0.046} & \textbf{0.028} & \textbf{0.018} & \textbf{0.037} & 64'600 \\
$L=256$ & 0.051 & 0.032 & 0.021 & 0.040 & 18'800 \\
\bottomrule
\end{tabular}
\end{table}

\subsubsection{Number of Contact Candidates}
\label{sec:num_cp_scaling_ablation}

Using the ground motions, we study the effect of changing the number of contact candidates. We compare the baseline configuration with 4 contact points (\figref{fig:handpicked_4}) to denser configurations of 10 and 18 hand-picked contact points distributed over the robot (Figs.~\ref{fig:handpicked_18}~and~\ref{fig:handpicked_10}). See \tabref{tab:ablation_comparison_contact_point_number_ground_motion} for results. 

For CoCo-InEKF, a positive correlation between contact point number and estimation accuracy is seen, at the expense of computational cost. The SET baseline shows limited improvement from 10 to 18 contacts.

\begin{table}[tb]
\caption{Ablation study on the scaling of the number of contact points. We report linear velocity ATE on synthetic ground motion data, number of parameters, and inference time for the neural network (NN) and the full state estimator (SE), respectively. For SET, a single value is reported, as the SE consists solely of the NN.}
\label{tab:ablation_comparison_contact_point_number_ground_motion}
\centering
\footnotesize
\begin{tabular}{lrrrrr}
\toprule
\textbf{Model}
 & $N$ & RMSE & \# Params. & NN / SE [ms] \\
\midrule
CoCo-InEKF  & 4 & 0.134 & 240’344 & 0.14 / 0.42 \\
CoCo-InEKF  & 10 & 0.099 & 334'844 & 0.18 / 0.87 \\
CoCo-InEKF  & 18 & \textbf{0.069} & 460'844 & 0.26 / 2.01 \\
SET, small & 10 & 0.107 & 815'363 & 2.03  \\
SET, large & 10 & 0.096 & 4'779'523 & 3.06 \\
SET, small & 18 & 0.104 & 821'507 & 2.08 \\
SET, large & 18 & 0.094 & 4'791'811 & 3.12 \\
\bottomrule
\end{tabular}
\end{table}

\subsubsection{Automated Contact Candidate Selection}
\label{sec:ablation_contact_point_configuration}

\begin{figure}[t]
    \centering
    \begin{subfigure}[b]{0.22\linewidth}
        \includegraphics[width=\textwidth]{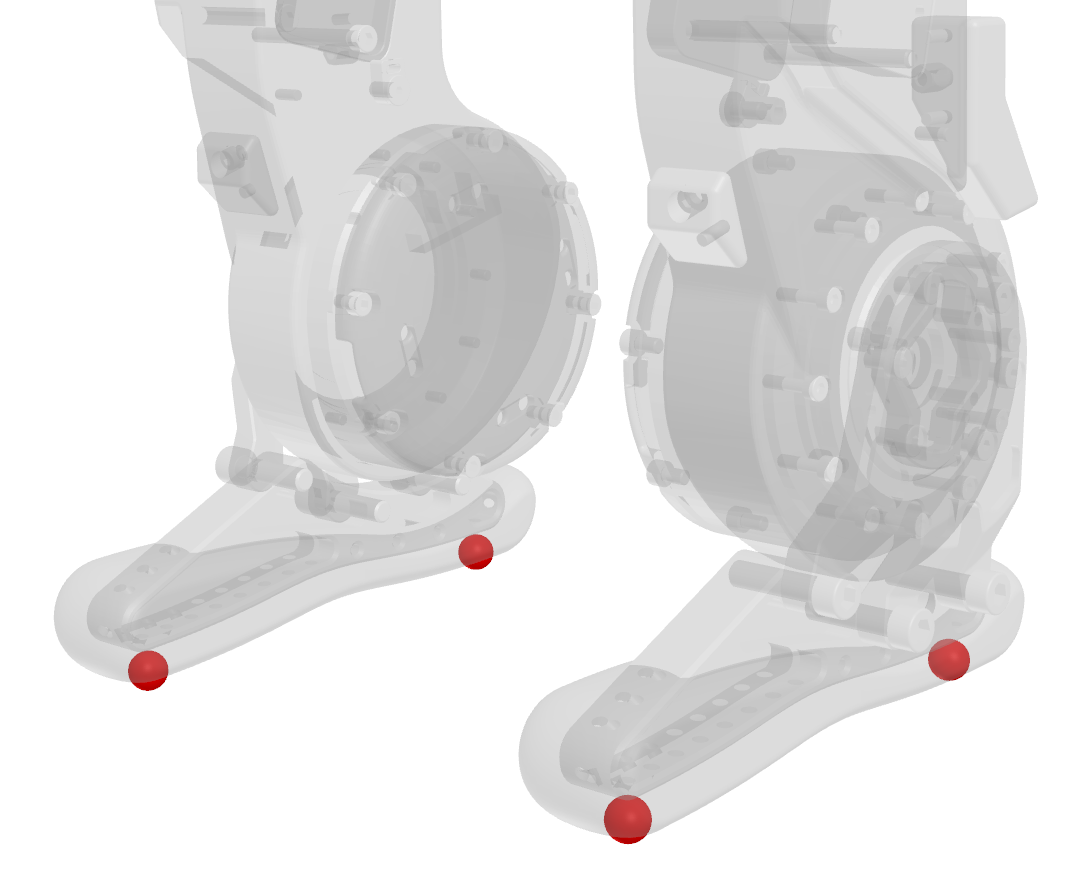}
        \caption{} \label{fig:handpicked_4}
    \end{subfigure}
    \hspace{0.01\linewidth}
    \begin{subfigure}[b]{0.22\linewidth}
        \includegraphics[width=\textwidth]{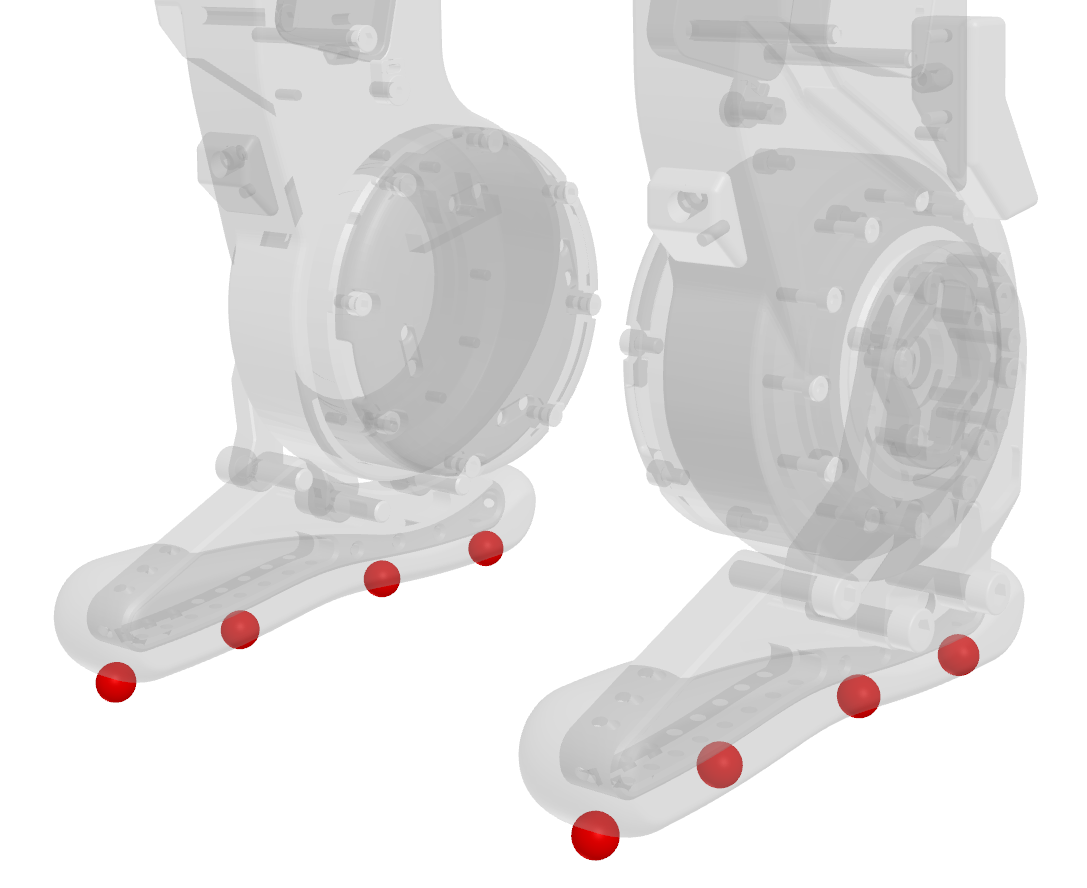}
        \caption{}
    \end{subfigure}
    \hspace{0.01\linewidth}
    \begin{subfigure}[b]{0.22\linewidth}
        \includegraphics[width=\textwidth]{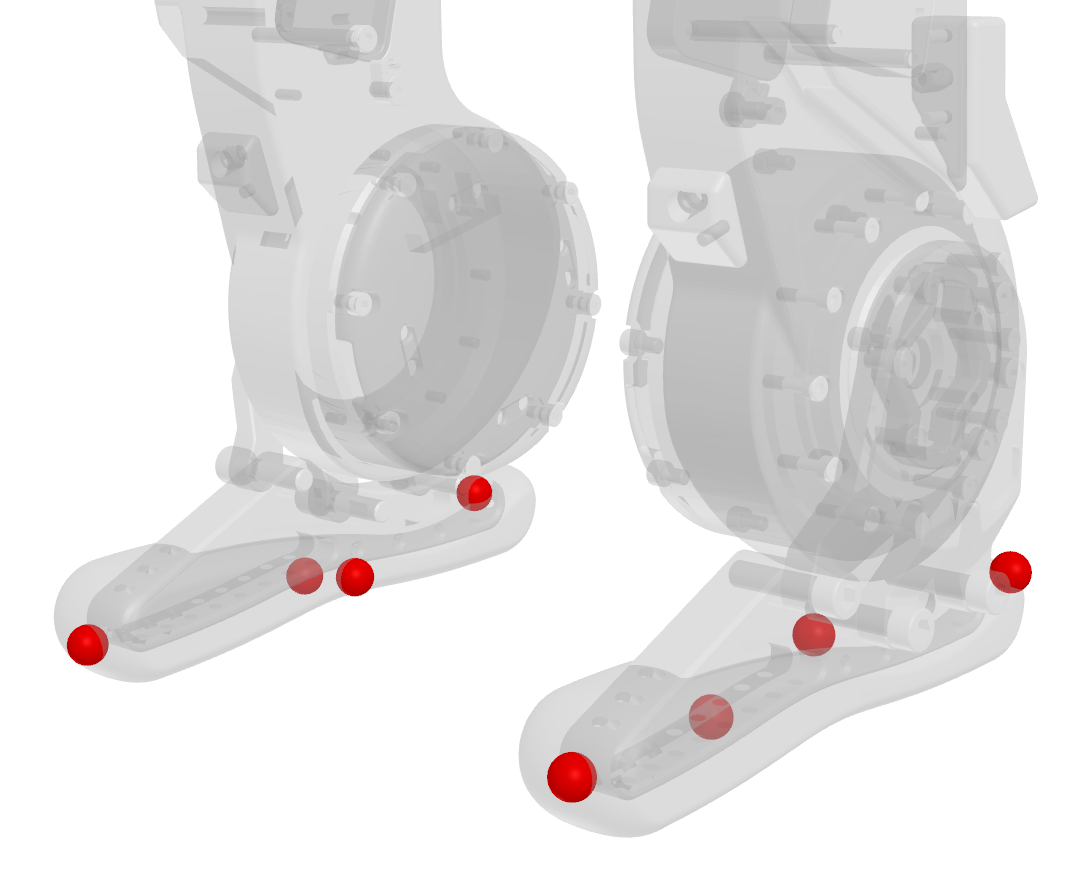}
        \caption{}
    \end{subfigure}
    \hspace{0.01\linewidth}
    \begin{subfigure}[b]{0.22\linewidth}
        \includegraphics[width=\textwidth]{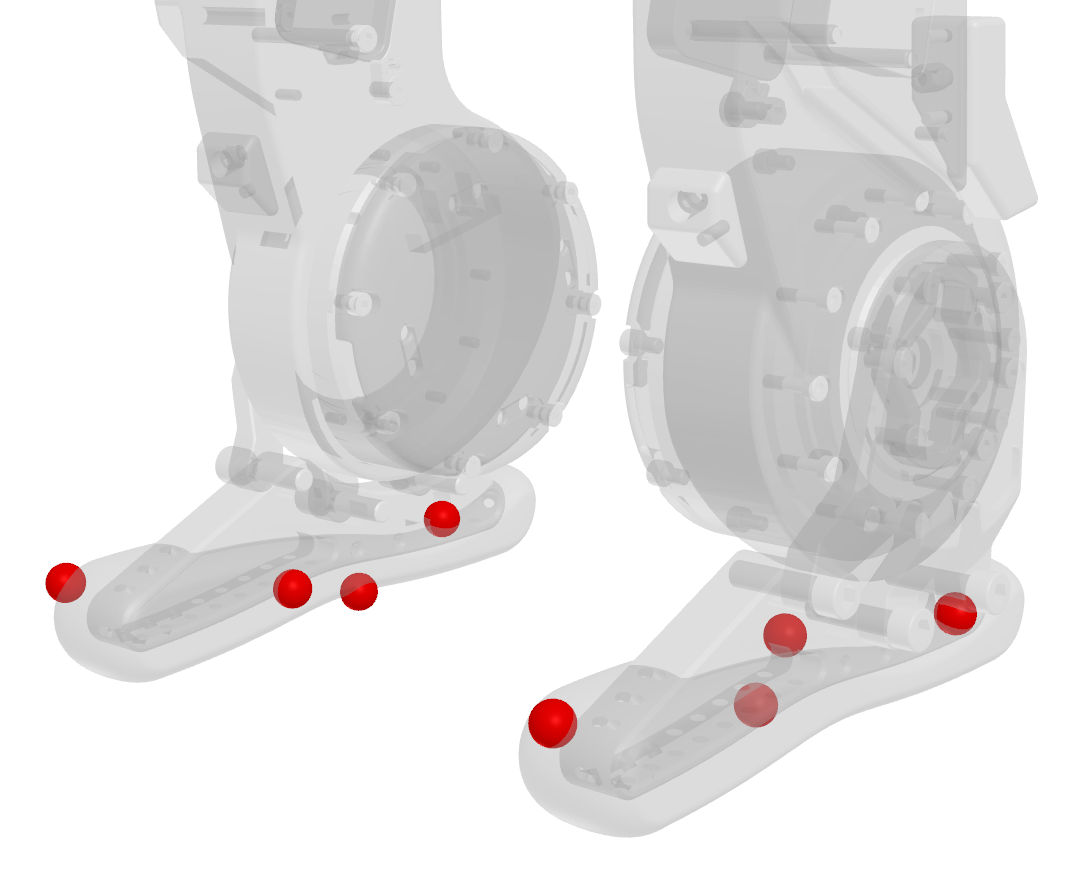}
        \caption{}
    \end{subfigure}
    \\[0.3cm]
    \begin{subfigure}[b]{0.22\linewidth}
        \includegraphics[width=\textwidth]{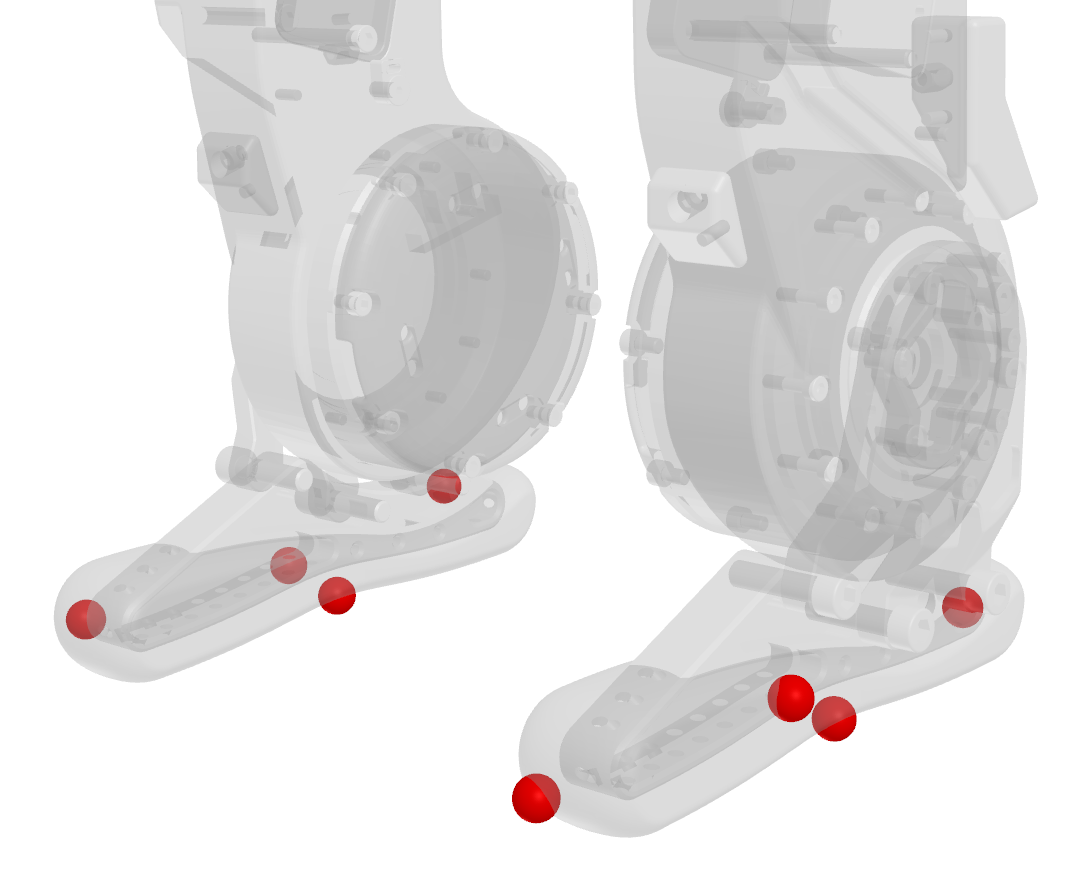}
        \caption{}
    \end{subfigure}
    \hspace{0.01\linewidth}
    \begin{subfigure}[b]{0.22\linewidth}
        \includegraphics[width=\textwidth]{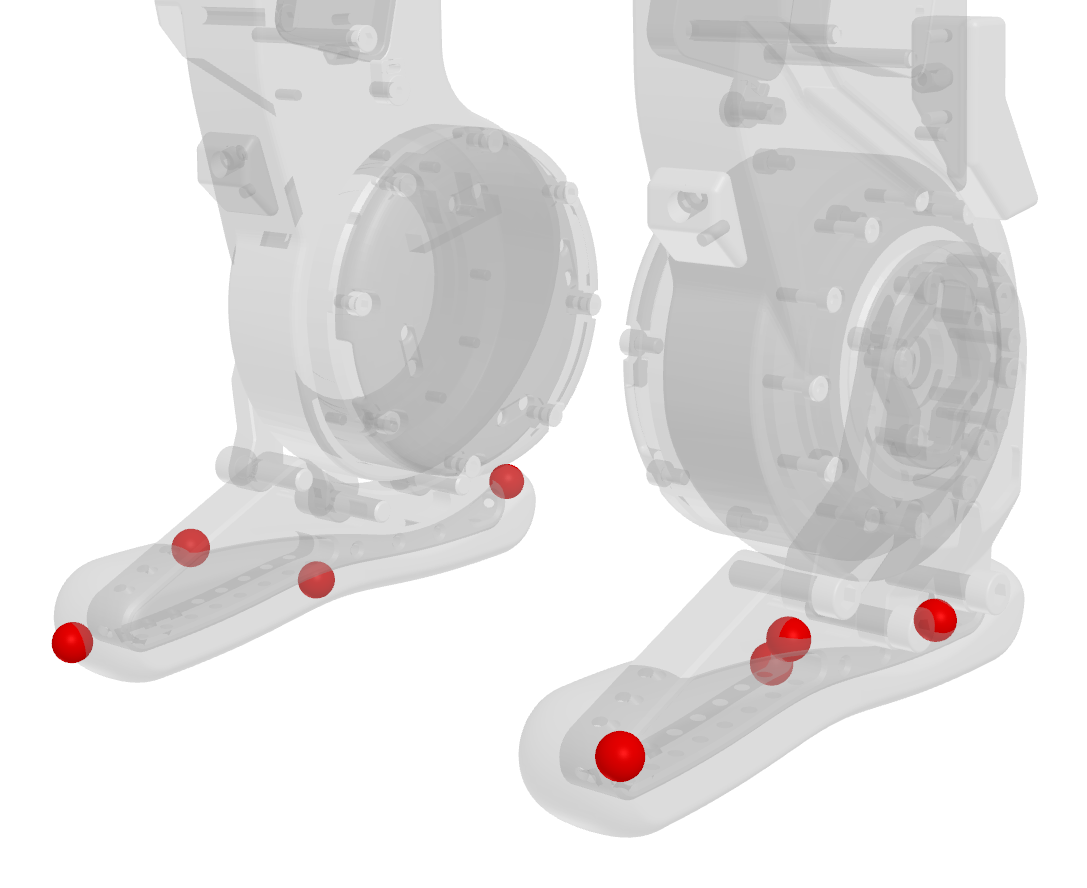}
        \caption{}
    \end{subfigure}
    \hspace{0.01\linewidth}
    \begin{subfigure}[b]{0.22\linewidth}
        \includegraphics[width=\textwidth]{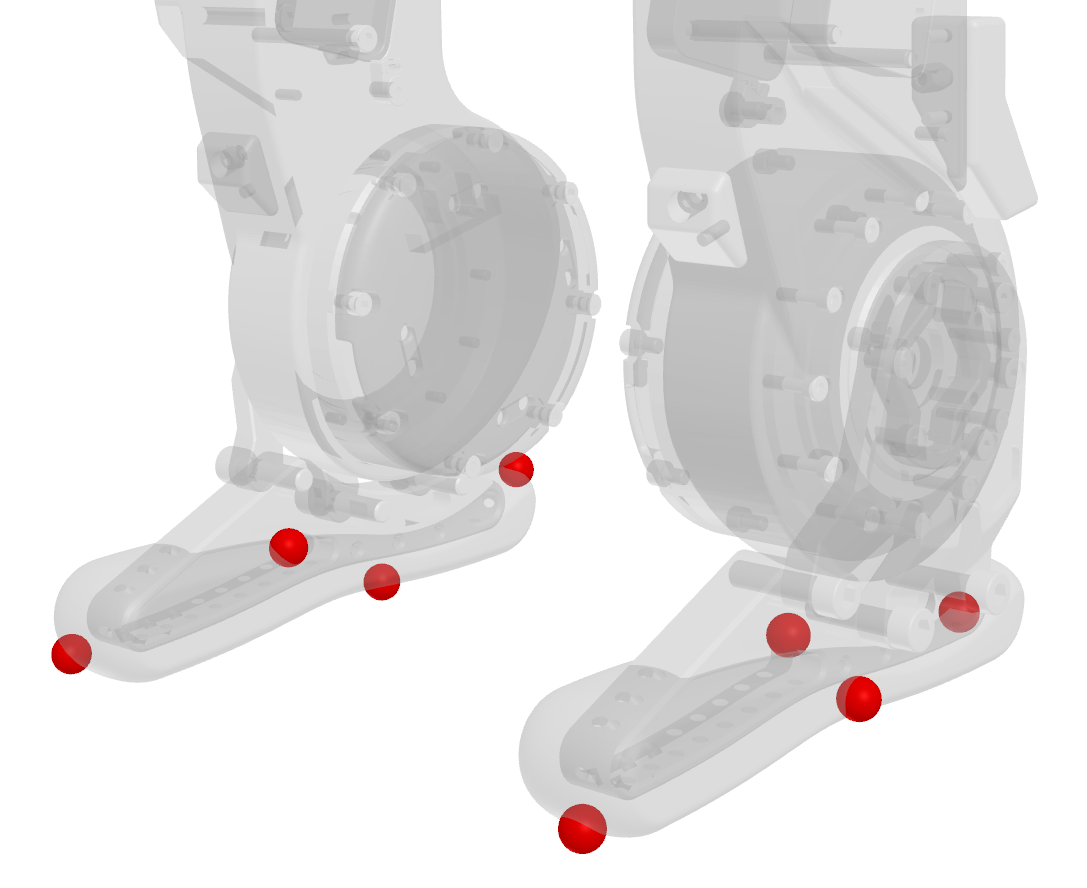}
        \caption{}
    \end{subfigure}
    \hspace{0.01\linewidth}
    \begin{subfigure}[b]{0.22\linewidth}
        \includegraphics[width=\textwidth]{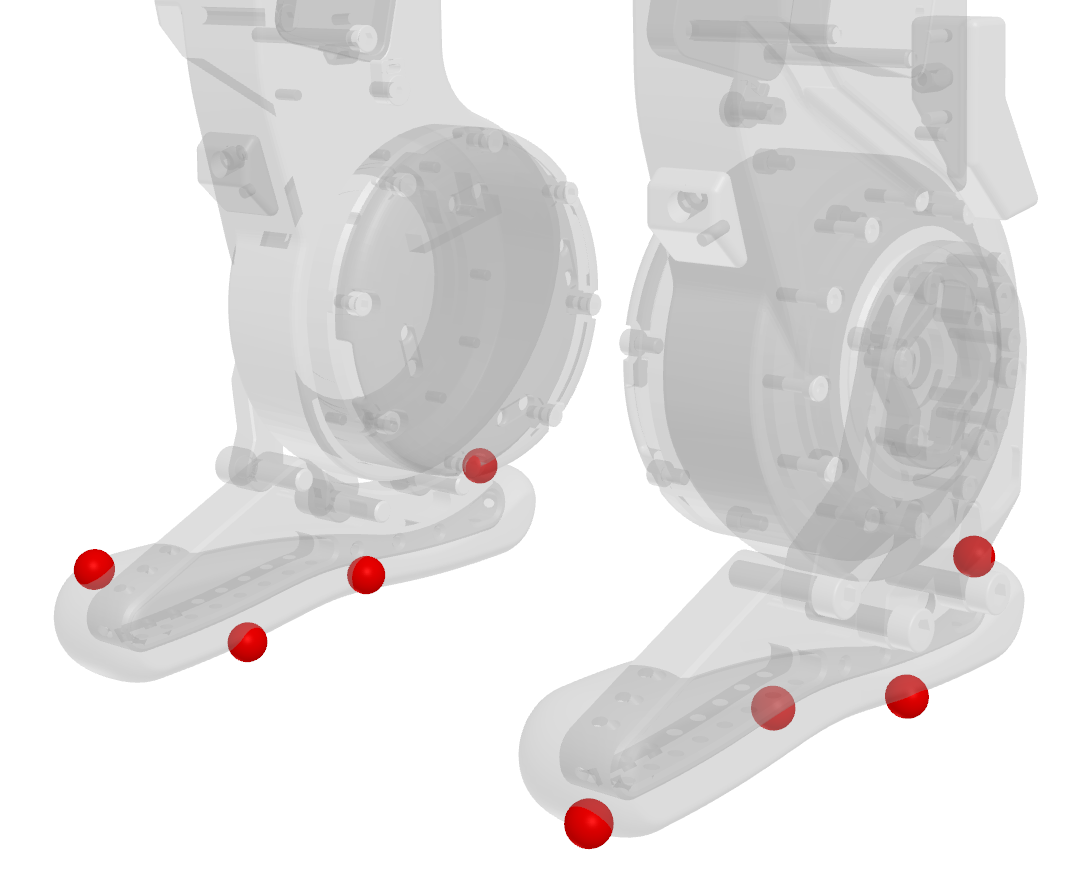}
        \caption{}
    \end{subfigure}
    \\[0.3cm]
    \begin{subfigure}[b]{0.22\linewidth}
        \includegraphics[width=\textwidth]{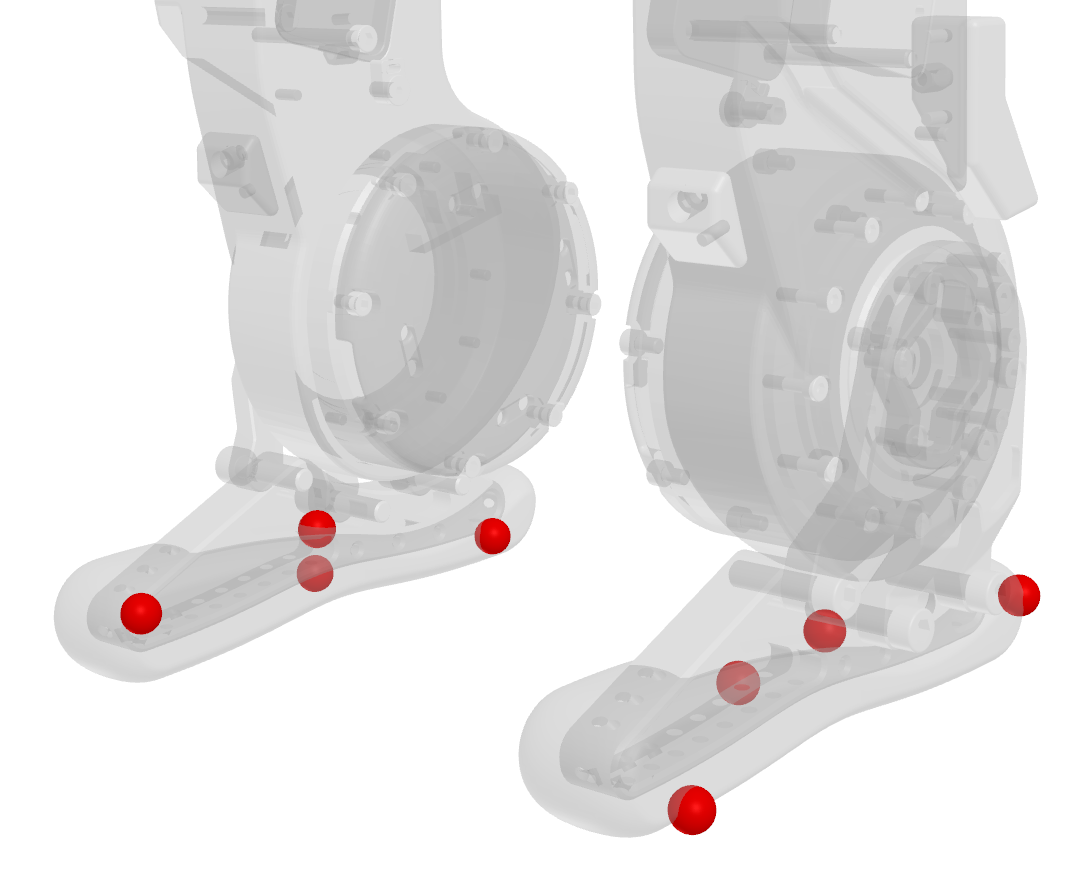}
        \caption{}
    \end{subfigure}
    \hspace{0.01\linewidth}
    \begin{subfigure}[b]{0.22\linewidth}
        \includegraphics[width=\textwidth]{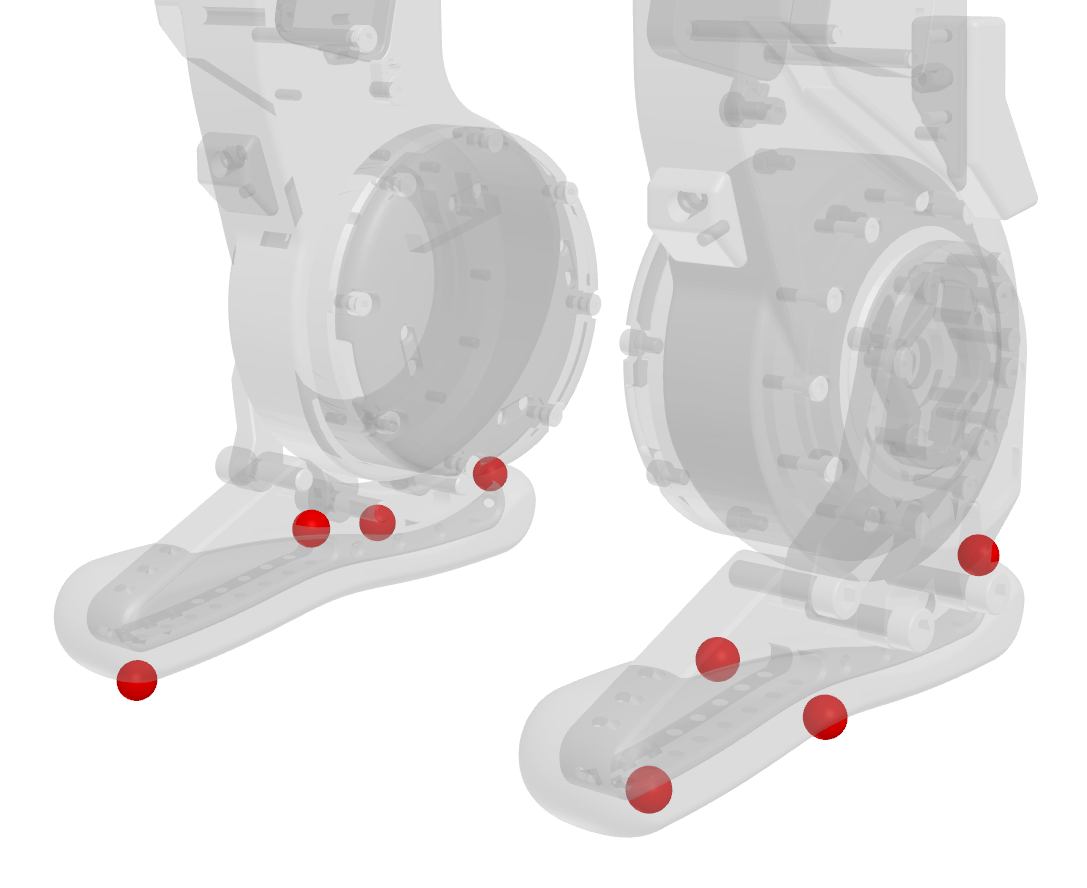}
        \caption{}
    \end{subfigure}
    \hspace{0.01\linewidth}
    \begin{subfigure}[b]{0.22\linewidth}
        \includegraphics[width=\textwidth]{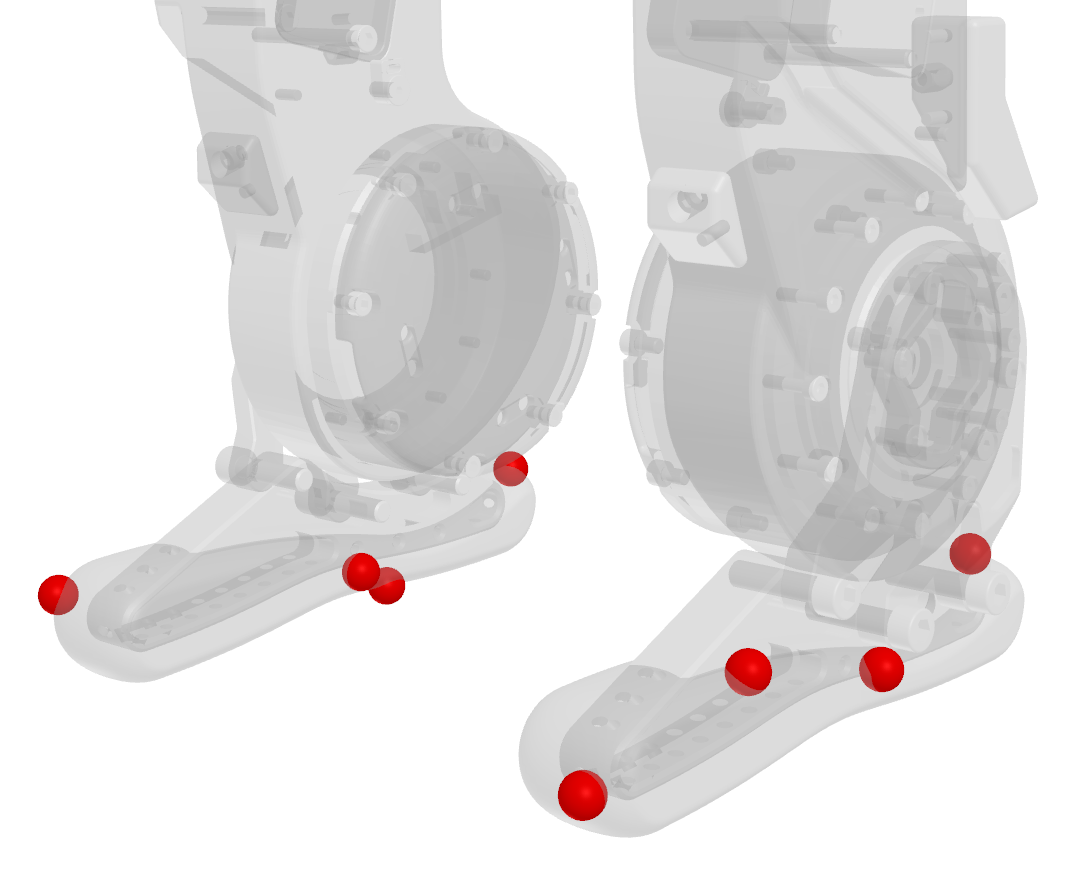}
        \caption{}
    \end{subfigure}
    \caption{\textbf{Foot contact candidate configurations.} As used in the evaluation and ablation study for dancing motions. Configurations (a) and (b), with 4 and 8 candidates respectively, were handpicked. All others have 8 candidates and were automatically generated using our proposed sampling method. }
    \label{fig:contact_configurations_feet}
\end{figure}

\begin{figure}[t]
    \centering
    \begin{subfigure}[b]{0.22\linewidth}
        \includegraphics[width=\textwidth]{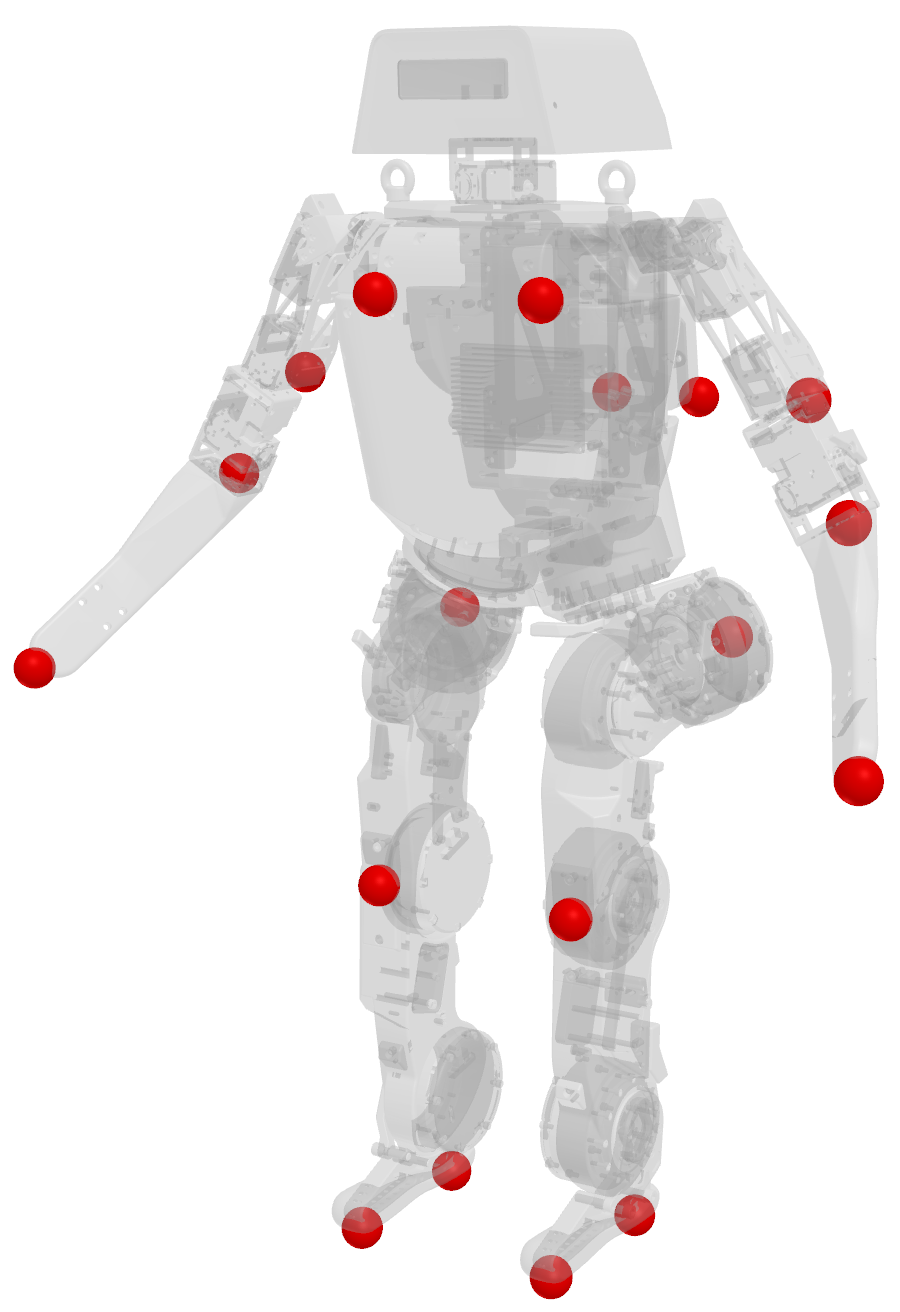}
        \caption{} \label{fig:handpicked_18}
    \end{subfigure}
    \hspace{0.01\linewidth}
    \begin{subfigure}[b]{0.22\linewidth}
        \includegraphics[width=\textwidth]{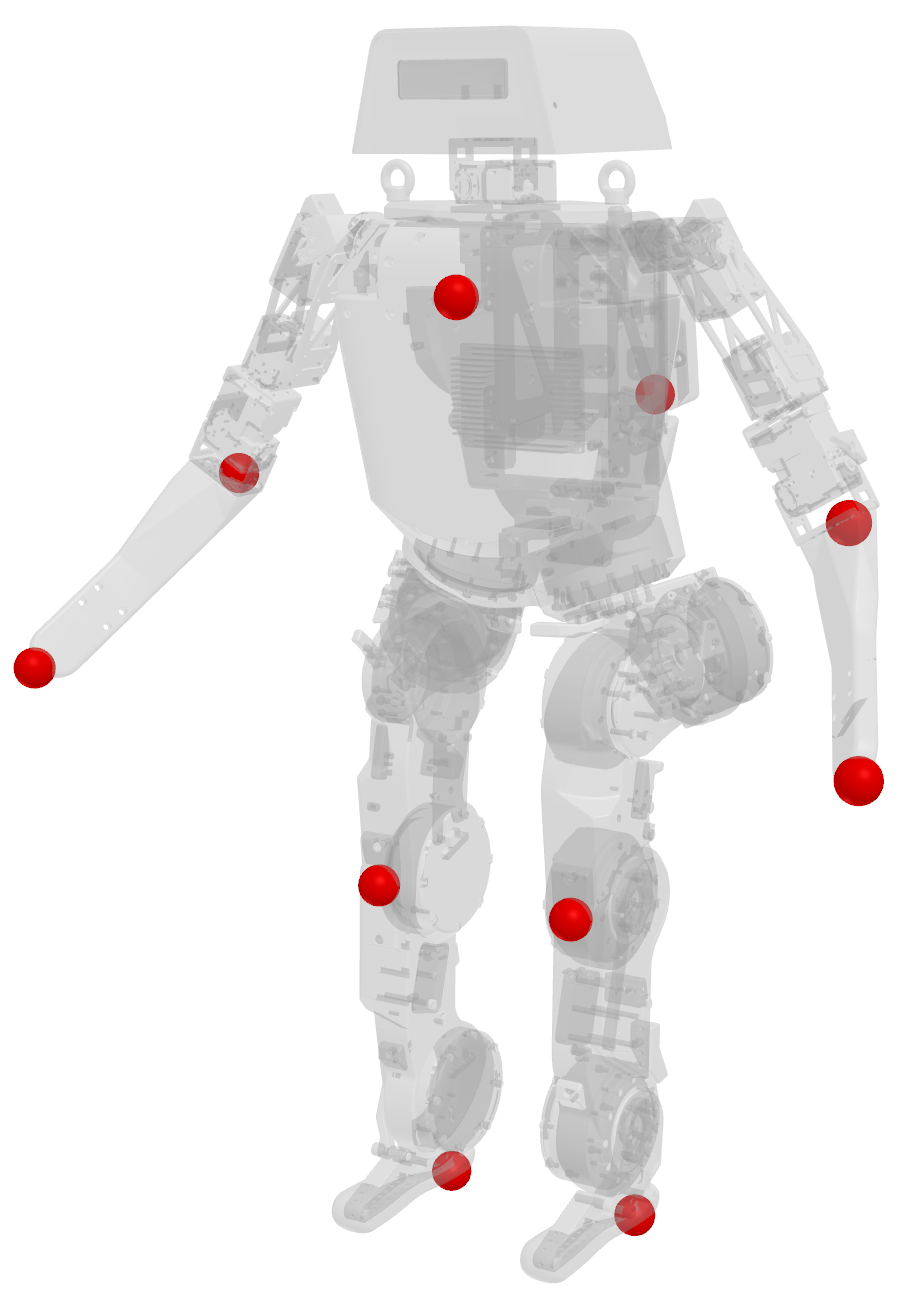}
        \caption{} \label{fig:handpicked_10}
    \end{subfigure}
    \hspace{0.01\linewidth}
    \begin{subfigure}[b]{0.22\linewidth}
        \includegraphics[width=\textwidth]{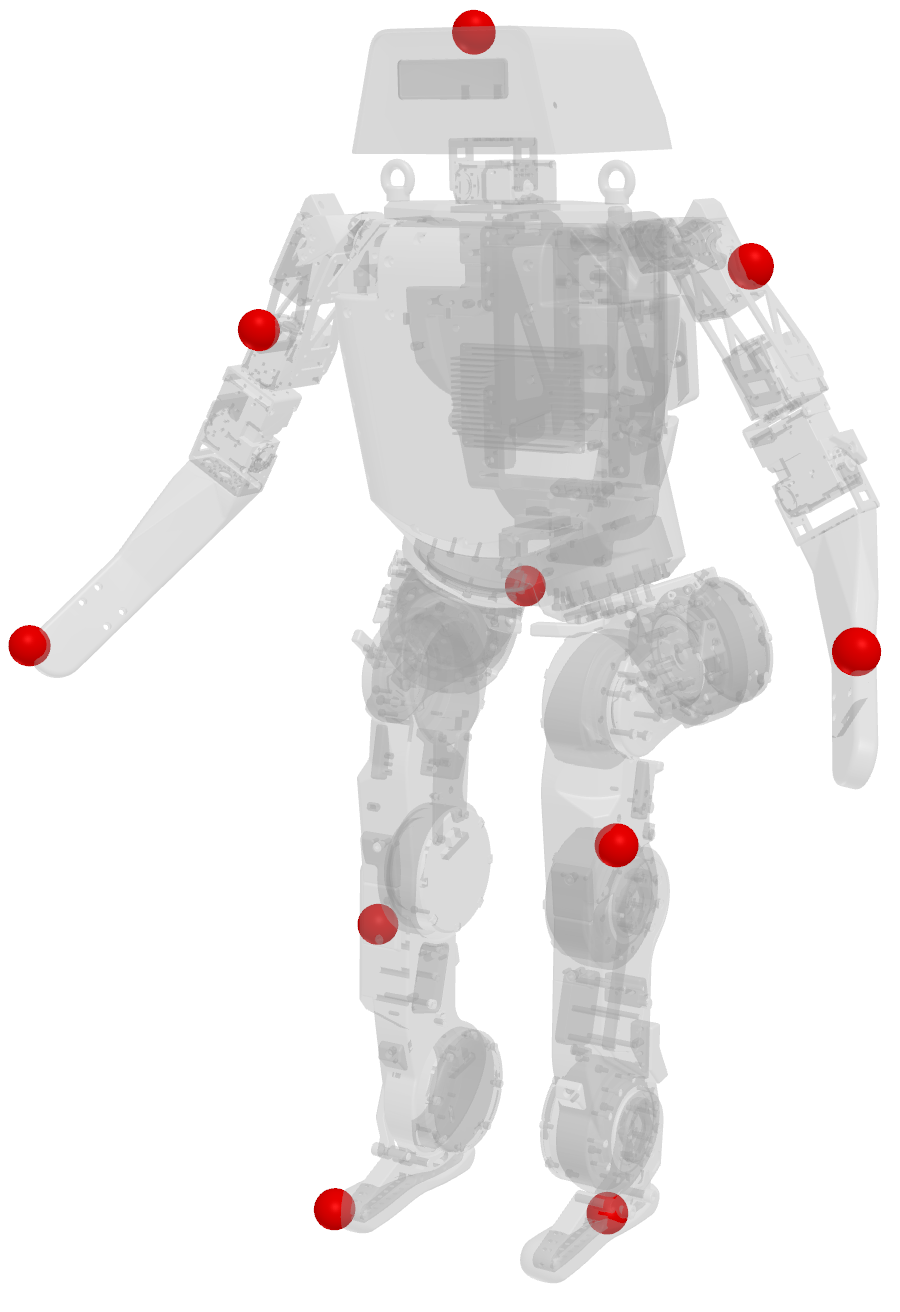}
        \caption{}
    \end{subfigure}
    \hspace{0.01\linewidth}
    \begin{subfigure}[b]{0.22\linewidth}
        \includegraphics[width=\textwidth]{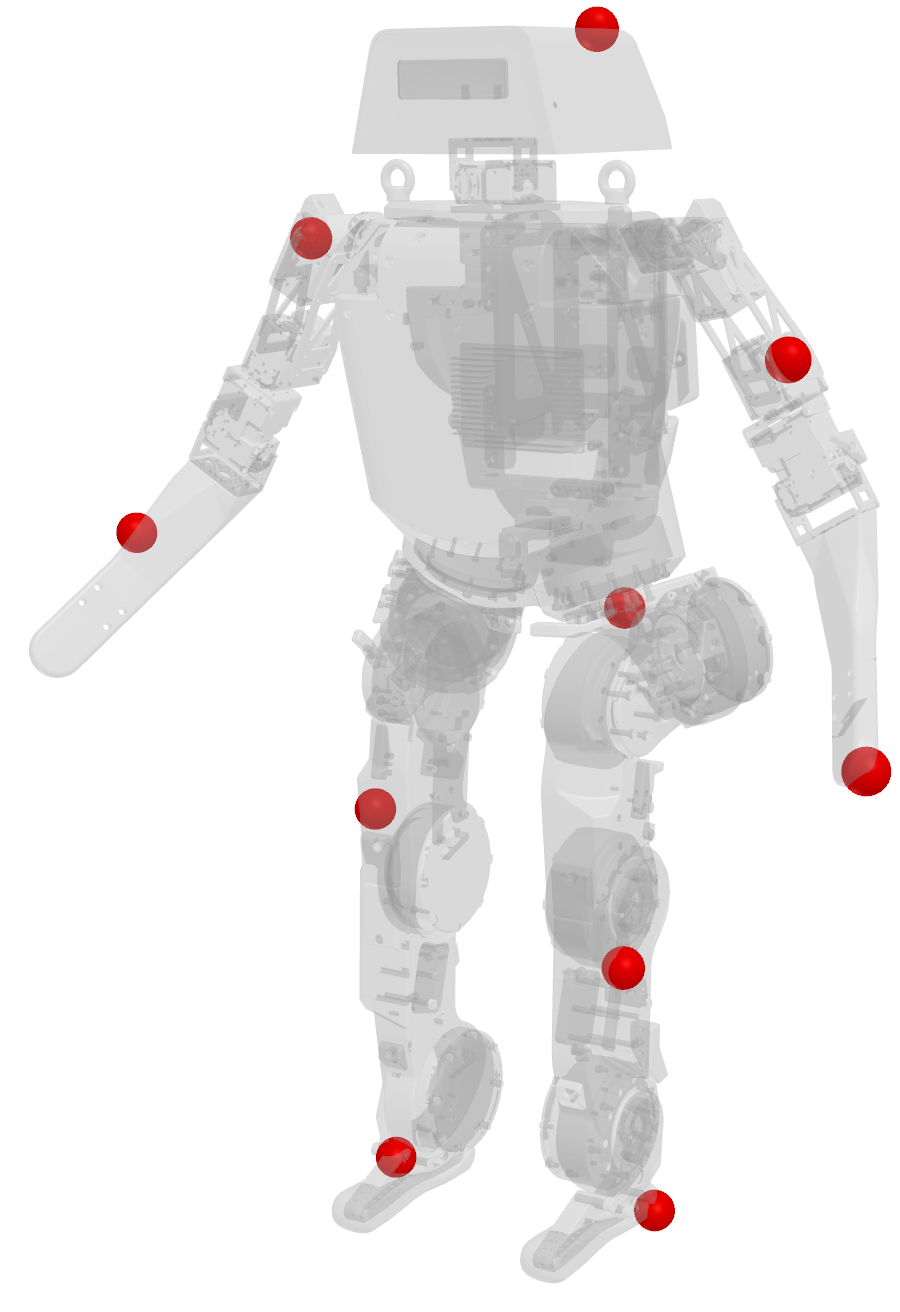}
        \caption{}
    \end{subfigure}
    \\[0.3cm]
    \begin{subfigure}[b]{0.22\linewidth}
        \includegraphics[width=\textwidth]{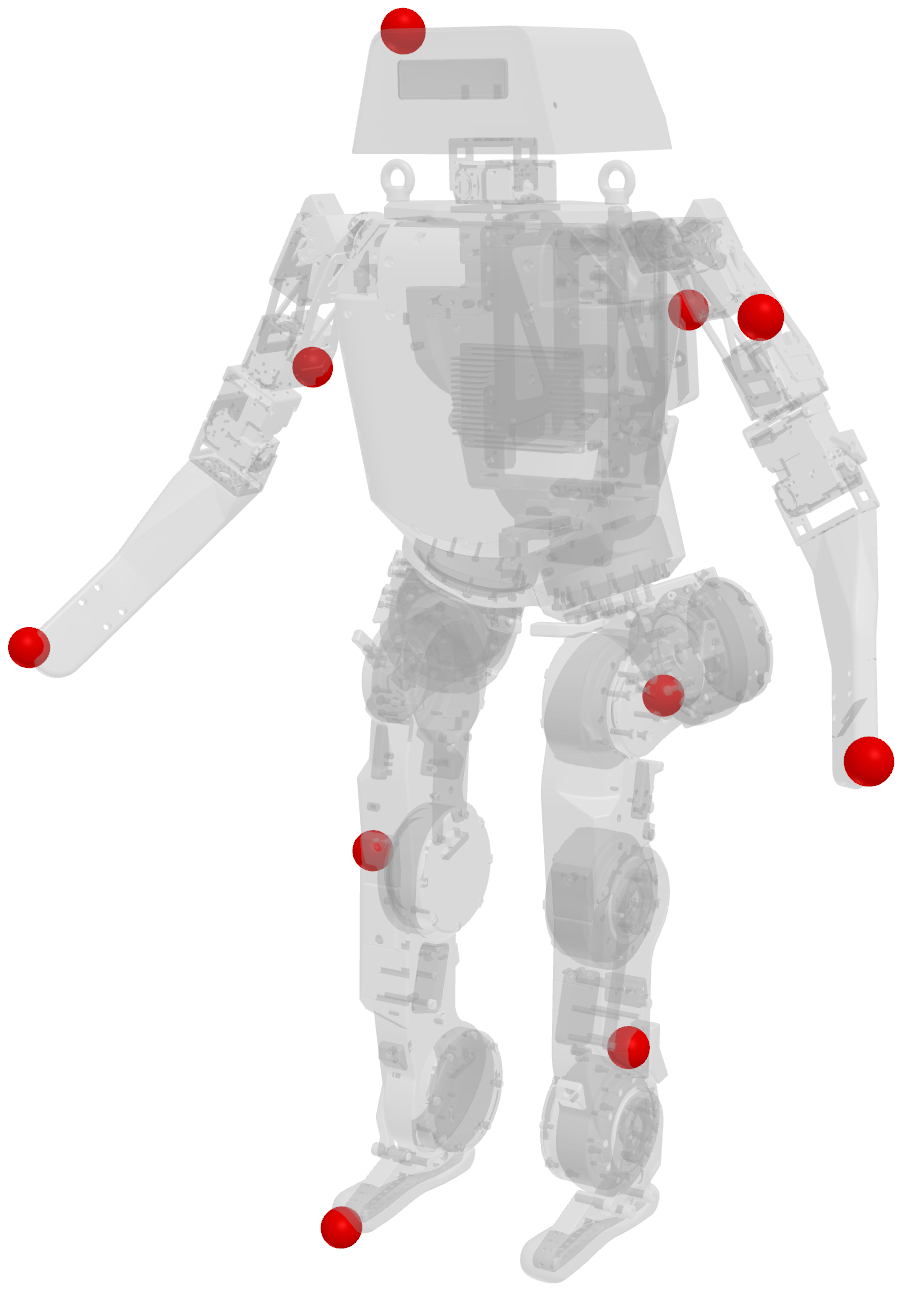}
        \caption{}
    \end{subfigure}
    \hspace{0.01\linewidth}
    \begin{subfigure}[b]{0.22\linewidth}
        \includegraphics[width=\textwidth]{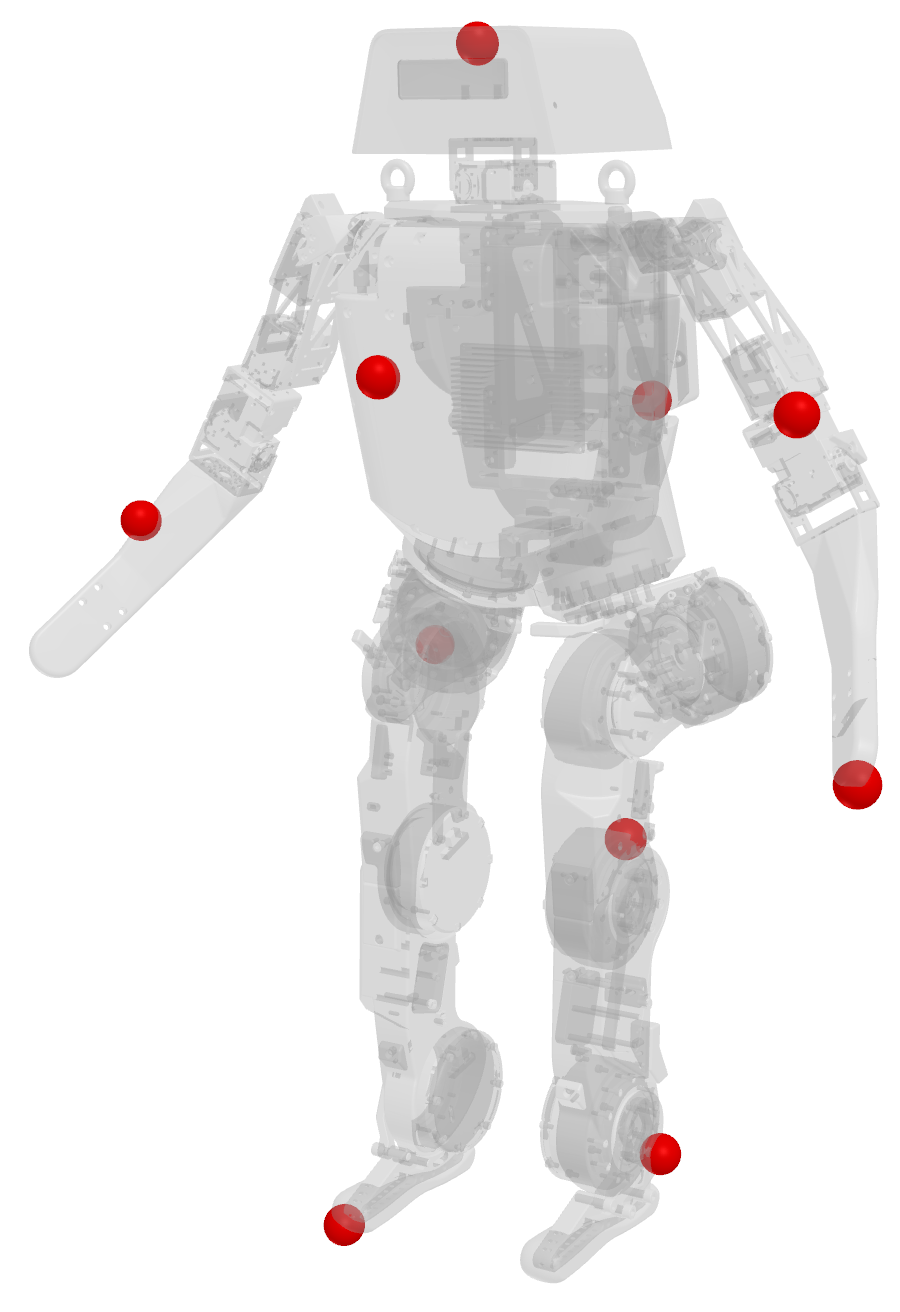}
        \caption{}
    \end{subfigure}
    \hspace{0.01\linewidth}
    \begin{subfigure}[b]{0.22\linewidth}
        \includegraphics[width=\textwidth]{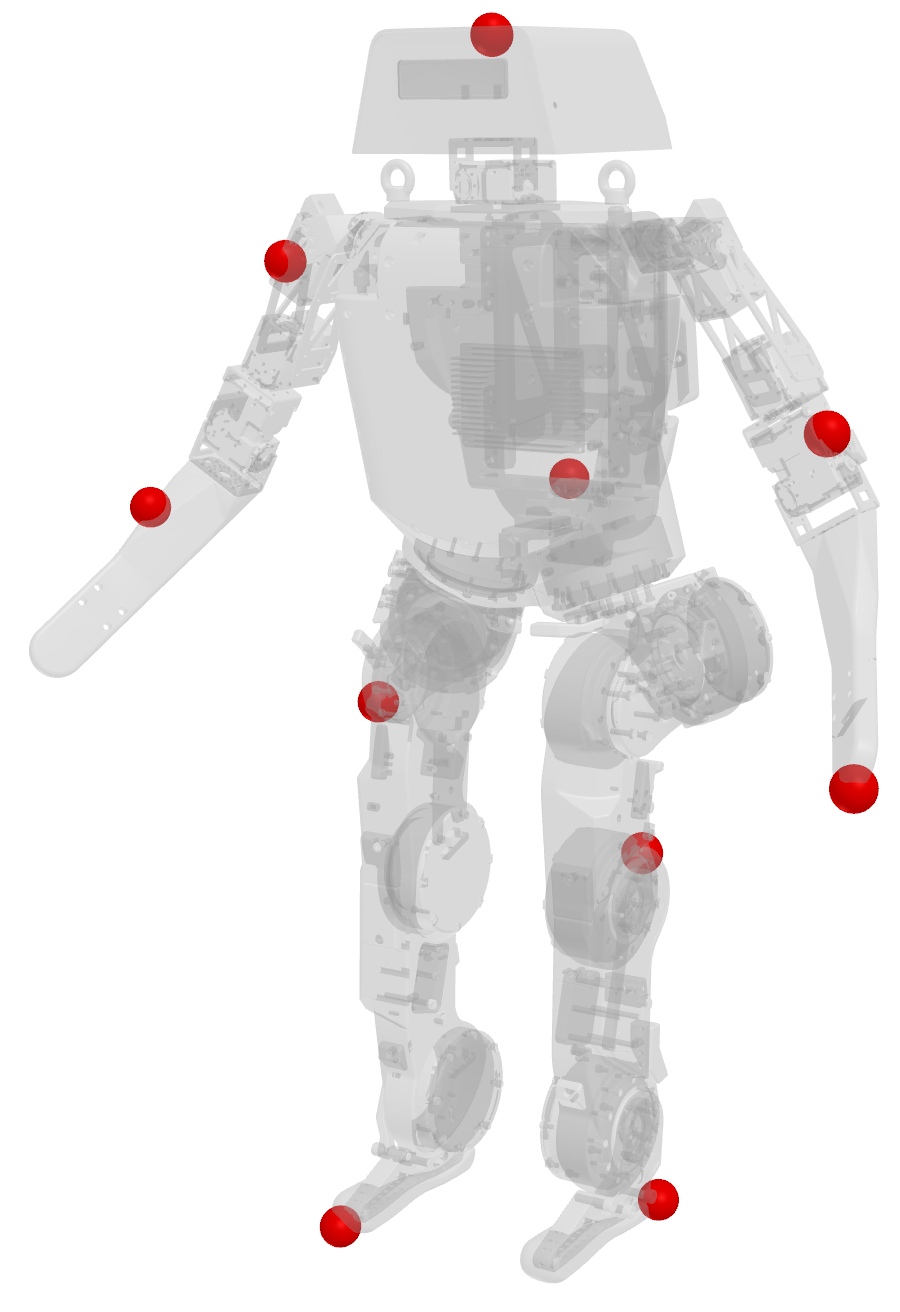}
        \caption{}
    \end{subfigure}
    \hspace{0.01\linewidth}
    \begin{subfigure}[b]{0.22\linewidth}
        \includegraphics[width=\textwidth]{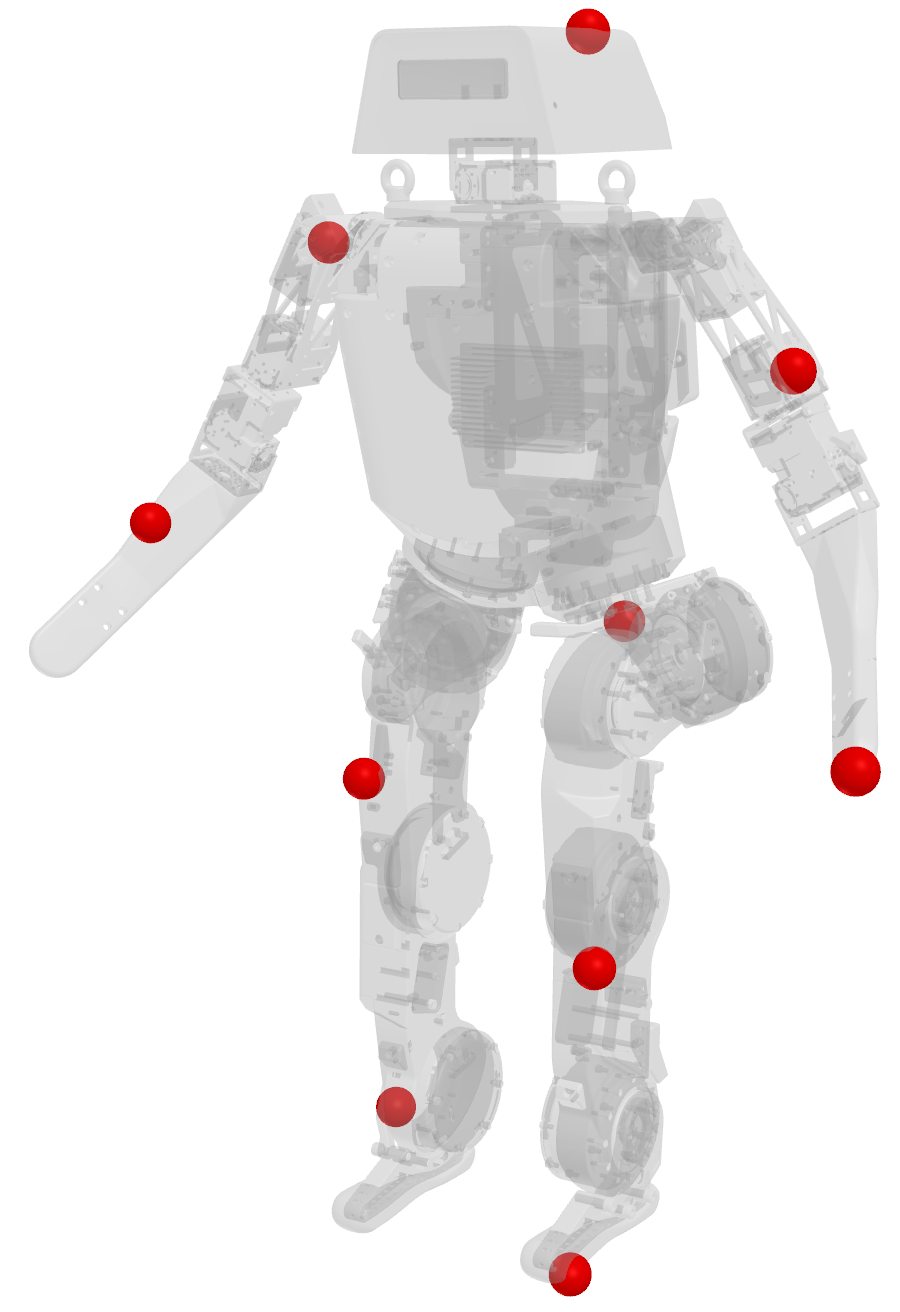}
        \caption{}
    \end{subfigure}
    \\[0.3cm]
    \begin{subfigure}[b]{0.22\linewidth}
        \includegraphics[width=\textwidth]{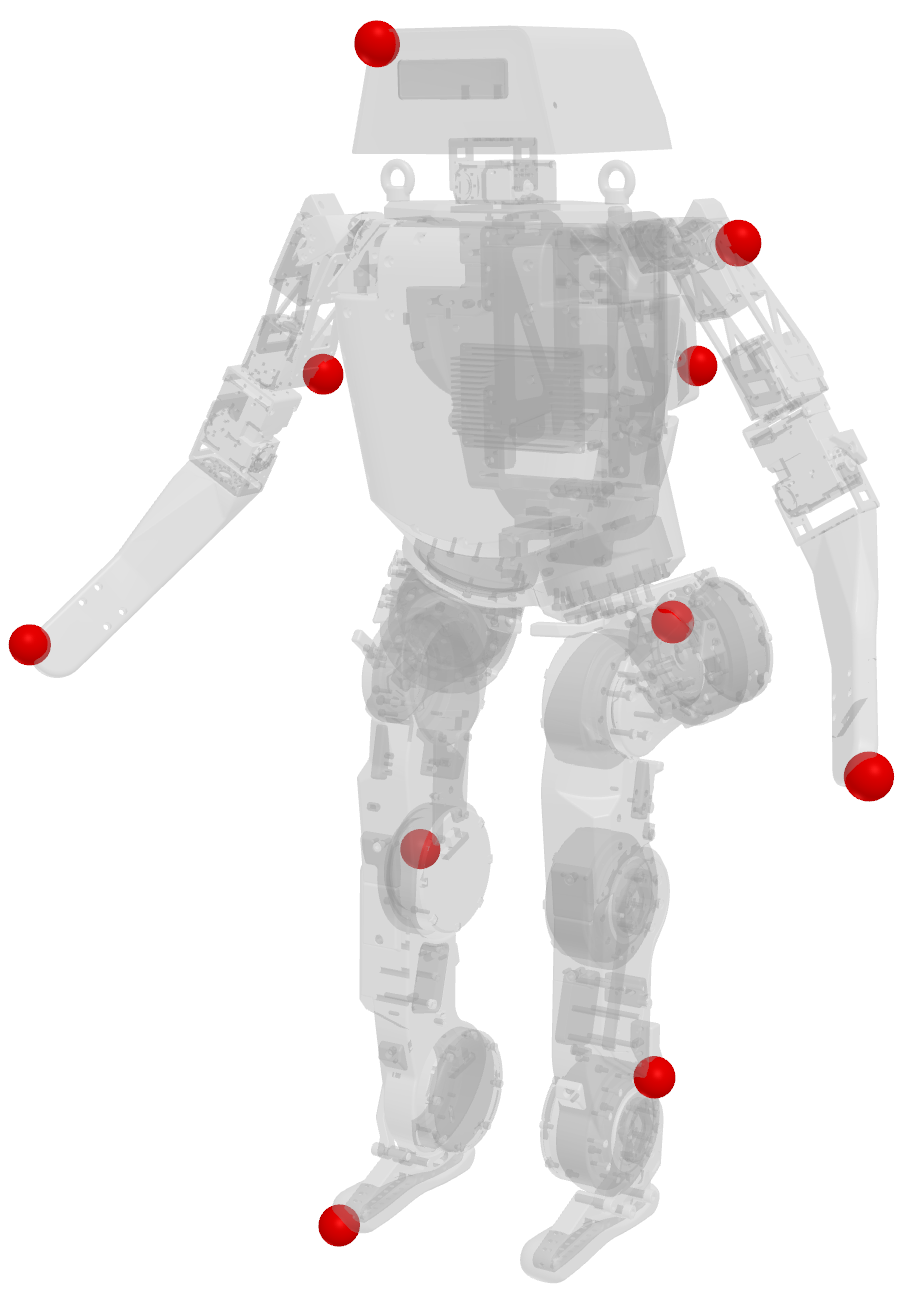}
        \caption{}
    \end{subfigure}
    \hspace{0.01\linewidth}
    \begin{subfigure}[b]{0.22\linewidth}
        \includegraphics[width=\textwidth]{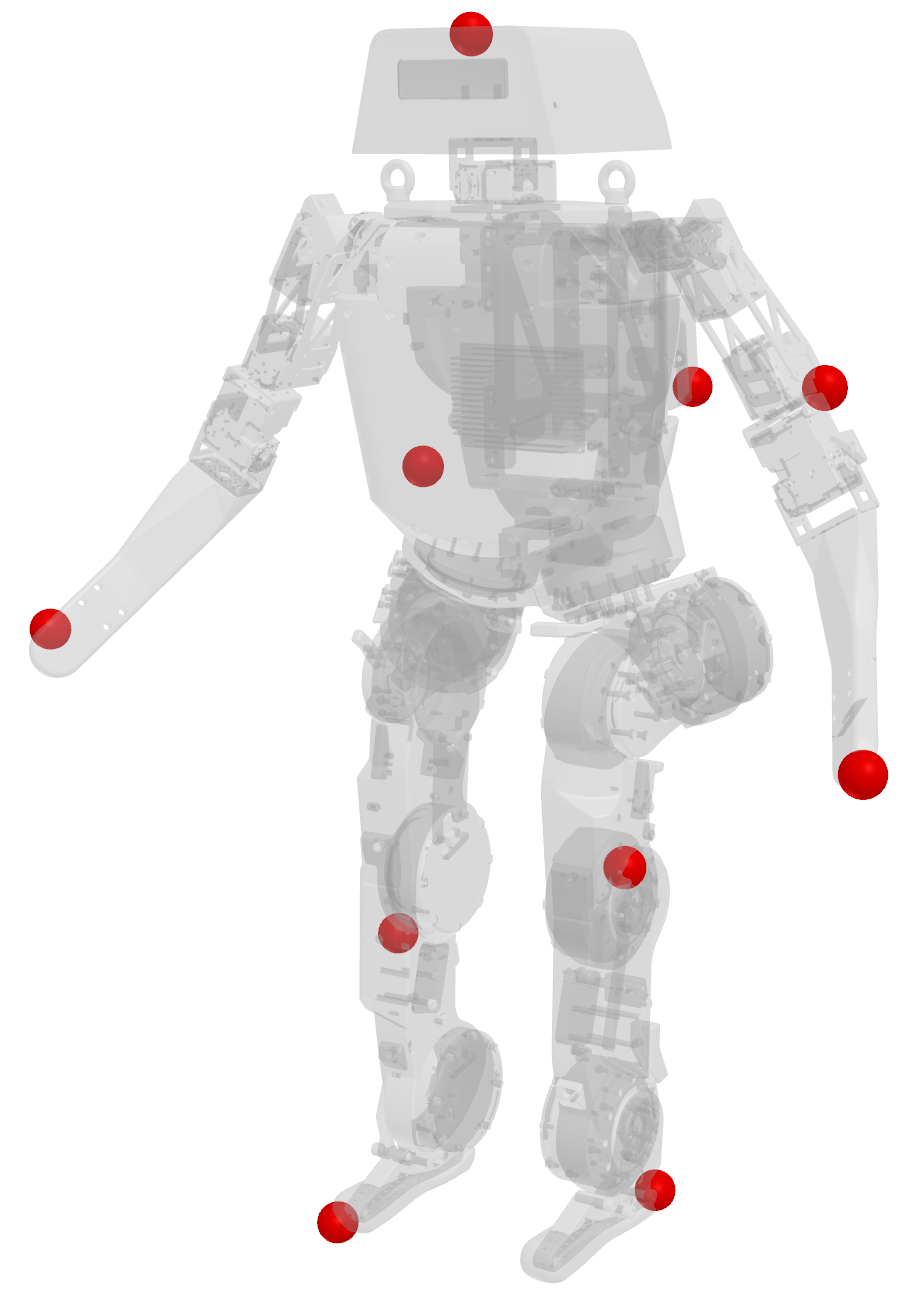}
        \caption{}
    \end{subfigure}
    \hspace{0.01\linewidth}
    \begin{subfigure}[b]{0.22\linewidth}
        \includegraphics[width=\textwidth]{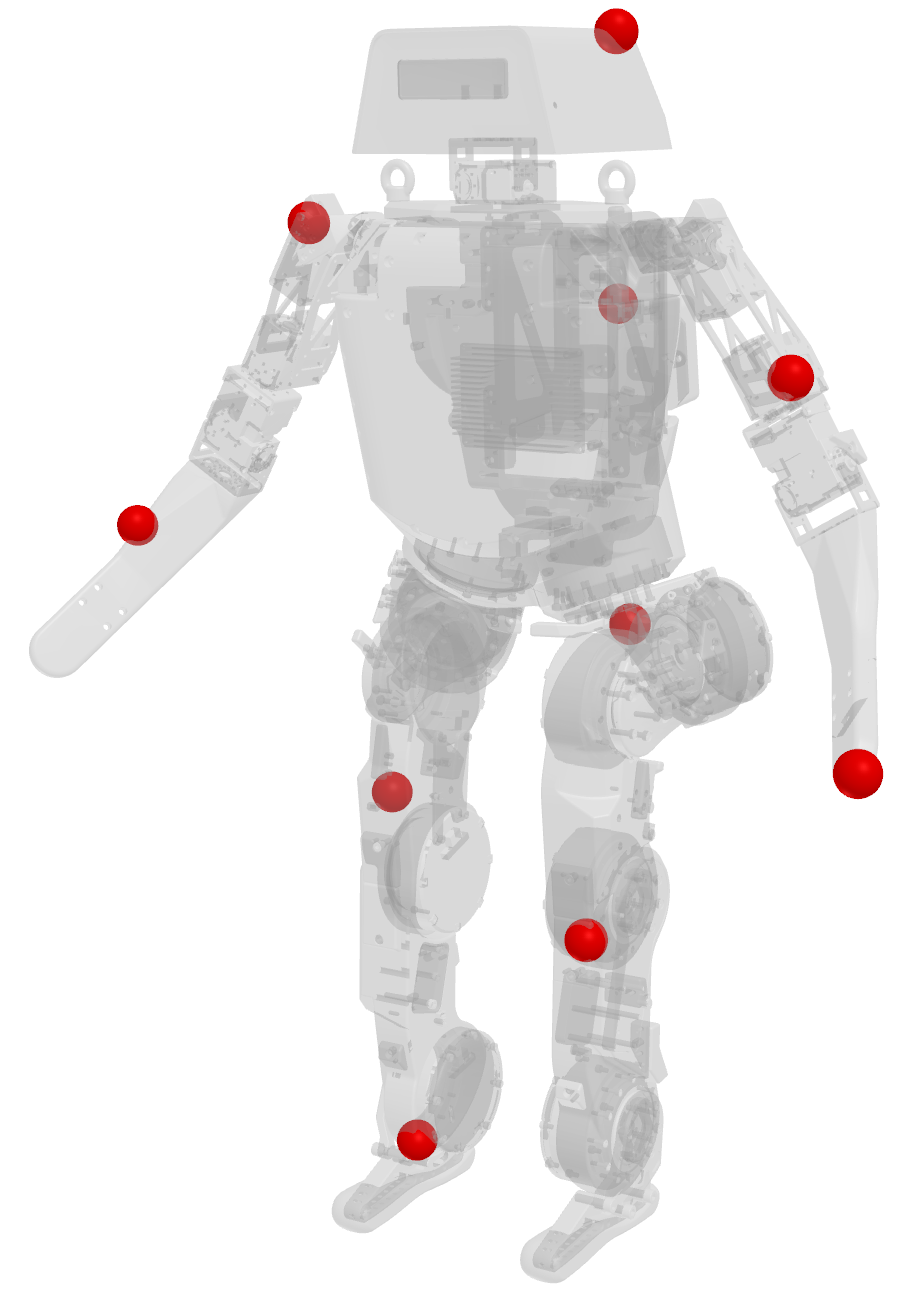}
        \caption{}
    \end{subfigure}
    \caption{\textbf{Full-body contact candidate configurations.} As used in the evaluation and ablation study for ground motions. Configuration (a) and (b), with 18 and 10 candidates respectively, were handpicked. All others have 10 candidates and were automatically generated using our proposed sampling method.}
    \label{fig:contact_configurations_full_body}
\end{figure}

To test our automated contact candidate selection approach, we consider two test cases: dancing motions ($N=8$, feet only), and ground motions ($N=10$, over the full body). For each test case, we hand-select a baseline and compare it to 9 samples from the automated method. The baselines and automated samples are shown in \figref{fig:contact_configurations_feet} for dancing motions and \figref{fig:contact_configurations_full_body} for ground motions.

Results are summarized in \tabref{tab:ablation_contact_point_location}. It can be seen that there is little variation across the randomly initialized, automated selections, and that the performance is similar to or better than the handpicked baseline. These results also support the hypothesis that the method is able to use the velocity covariance contact representation to reason about contact points that are not exactly in contact. 

\begin{table}
    \caption{The automated contact candidate selection compared to the handpicked baseline, for linear velocity ATE. Ranges indicate [worst, best] sample.}
    \label{tab:ablation_contact_point_location}
    \centering
    \footnotesize
    \begin{tabular}{l r r}
    \toprule
     & RMSE \\
     \midrule
     Dancing motions, automated & [0.056, 0.052] \\
     Dancing motions, handpicked & 0.057 \\
     Ground motions, automated & [0.104, 0.092] \\
     Ground motions, handpicked & 0.099 \\
     \bottomrule
    \end{tabular}
\end{table}

\subsection{Statistical Filter Consistency}

\begin{figure}[tbp]
  \centering
  \includegraphics[width=\linewidth]{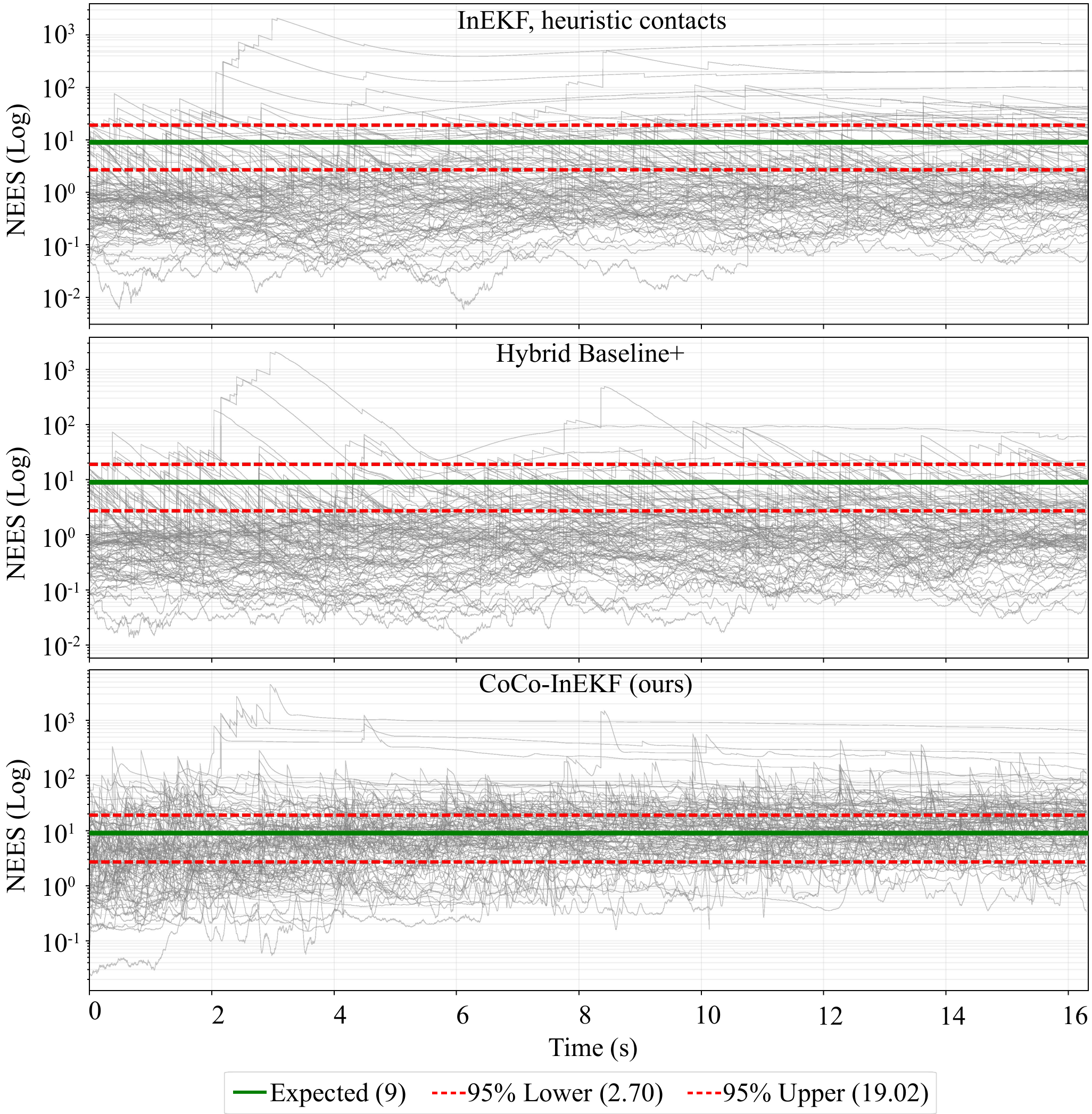}
  \caption{\textbf{Consistency.} Visualization of the normalized estimation error squared (NEES) of the combined core state of baseline InEKF approaches vs. our proposed CoCo-InEKF for 100 dance motion sequences. CoCo-InEKF's NEES values are more consistent with the expected 95\% confidence interval of a $\chi^2$ distribution. 
  We omit the Hybrid Baseline, as it performs near identical to the Hybrid Baseline+.
  }
  \label{fig:nees_pairwise_eval}
\end{figure}

To assess the consistency of the proposed filter architecture, we employ the
normalized estimation error squared (NEES) metric~\cite{BarShalom2001}. A state
estimator is called consistent if it is unbiased and if the actual estimation
error covariance matches the covariance reported by the filter at that time step, i.e.,
\begin{equation}
    \mathbb{E}\!\left[\tilde{\mathbf{x}}\right] = \mathbf{0}, \qquad
    \mathbb{E}\!\left[\tilde{\mathbf{x}}\,\tilde{\mathbf{x}}^{\top}\right]
    = \mathbf{P},
\end{equation}
with $\tilde{\mathbf{x}} = \mathbf{x} - \hat{\mathbf{x}}$, where the hat symbol ($\hat{\cdot}$) indicates the ground-truth state. The NEES normalizes the estimation error by the filter's covariance,
\begin{equation}
    \epsilon = \tilde{\mathbf{x}}^{\top}\,\mathbf{P}^{-1}\,
    \tilde{\mathbf{x}},
\end{equation}
and is, under the Gaussian assumption, $\chi^{2}$-distributed with $n_x = \dim(\tilde{\mathbf{x}})$
degrees of freedom, yielding $\mathbb{E}[\epsilon] = n_x$. Consistency is then evaluated by checking whether $\epsilon$
lies within the two-sided $(1-\alpha)$ confidence interval $[r_1, r_2]$ obtained from the inverse $\chi^{2}$ cumulative distribution function. Values exceeding $r_2$ indicate an optimistic (overconfident) filter, whereas values below $r_1$
indicate a conservative (pessimistic) one. For a comprehensive treatment, the reader is referred to~\cite{BarShalom2001}.

\begin{table}[tb]
\caption{NEES evaluation on the simulated dancing test data, reported as the percentage of time steps with NEES values within the 95\% confidence bounds ($[2.7, 19]$ for the combined core state, $[0.22, 9.35]$ for the individual states).}
\label{tab:nees_comparison_dancing}
\centering
\footnotesize
\setlength{\tabcolsep}{4pt}
\begin{tabular}{lrrrr}
\toprule
\textbf{Model} & \textbf{Core} & \textbf{Vel.} & \textbf{Pos.} & \textbf{Ori.} \\
\midrule
InEKF, GT contacts        & 37.7\% & 66.8\% & \textbf{60.7\%} & 65.4\% \\
InEKF, heur.\ contacts    & 20.3\% & 47.4\% & 23.6\%          & 47.4\% \\
Hybrid Baseline           & 18.2\% & 50.2\% & 11.3\%          & 52.2\% \\
Hybrid Baseline+          & 18.4\% & 50.3\% &  7.9\%          & 51.8\% \\
CoCo-InEKF (ours)         & \textbf{52.1\%} & \textbf{71.1\%} & 59.8\% & \textbf{68.1\%} \\
\bottomrule
\end{tabular}
\end{table}

All consistency evaluations are conducted on our diverse dance motion test set, consisting of 100 unique motion trajectories. As shown in \tabref{tab:nees_comparison_dancing}, CoCo-InEKF matches or exceeds the consistency of the original formulation utilizing ground-truth, privileged information. The baseline methods that estimate the contact states either heuristically or with a learned binary contact classification show lower overall consistency. They are underconfident across most of the evaluated trajectories, as evident in \figref{fig:nees_pairwise_eval}, whereas CoCo-InEKF exhibits NEES values more in line with the 95\% confidence interval extracted from the inverse $\chi^2$ cumulative distribution function. Our formulation thus improves the filter consistency, despite not explicitly optimizing for this metric.

\subsection{Real-World Experiments}

\begin{table*}[tb]
\caption{ATE comparison on 20 real-world ground motion sequences.}
\label{tab:ate_total_comparison_ground_motion_real}
\centering
\footnotesize
\begin{tabular}{lrrrrrrrrrrrr}
\toprule
 & \multicolumn{4}{c}{\textbf{Linear Velocity ATE}} & \multicolumn{4}{c}{\textbf{Position ATE}} & \multicolumn{4}{c}{\textbf{Orientation ATE}} \\
\cmidrule(lr){2-\numexpr5\relax} \cmidrule(lr){6-\numexpr9\relax} \cmidrule(lr){10-\numexpr13\relax}
\textbf{Model} & RMSE & MAE & MED & STD & RMSE & MAE & MED & STD & RMSE & MAE & MED & STD \\
 \midrule
InEKF, heuristic contacts & 1.5419 & 0.5305 & 0.0261 & 1.4478 & 21.3841 & 6.0165 & \textbf{0.0671} & 20.5203 & \textbf{0.0152} & \textbf{0.0110} & \textbf{0.0081} & 0.0105 \\
Hybrid Baseline* & 0.1178 & 0.0738 & 0.0453 & 0.0918 & 0.7212 & 0.3764 & 0.2093 & 0.6152 & 0.0152 & 0.0110 & 0.0085 & \textbf{0.0105} \\
Hybrid Baseline+ & 0.1699 & 0.1108 & 0.0669 & 0.1288 & 1.4869 & 0.7305 & 0.3396 & 1.2951 & 0.0188 & 0.0129 & 0.0095 & 0.0137 \\
CoCo-InEKF (ours) & \textbf{0.0805} & \textbf{0.0398} & \textbf{0.0167} & \textbf{0.0699} & 0.2019 & 0.1497 & 0.1181 & 0.1355 & 0.0302 & 0.0201 & 0.0130 & 0.0225 \\
SET, small* & 0.1002 & 0.0501 & 0.0193 & 0.0867 & \textbf{0.1665} & \textbf{0.1240} & 0.0952 & \textbf{0.1110} & 0.0257 & 0.0165 & 0.0109 & 0.0197 \\
SET, large* & 0.0974 & 0.0475 & 0.0174 & 0.0850 & 0.2245 & 0.1803 & 0.1498 & 0.1337 & 0.0257 & 0.0165 & 0.0109 & 0.0197 \\
\bottomrule
\end{tabular}
\end{table*}

\begin{table}[tb]
\caption{Success rate [\SI{}{\percent}] of various real-world dance motions with the state estimators in-the-loop. The pirouette and moonwalk are not part of the training set.}
\label{tab:success_chance_dancing}
\centering
\footnotesize
\begin{tabular}{lrrr}
\toprule

 & Dances & Pirouette & Moonwalk \\
\textbf{Model}
 &(training set) & (unseen) & (unseen) \\
\midrule
MoCap & 92 & 90 & \textbf{100} \\
InEKF, heuristic contacts & 77 & 60 & \textbf{100} \\
Hybrid Baseline+ & 85 & 50 & 10 \\
CoCo-InEKF (ours) & \textbf{95} & \textbf{100} & \textbf{100} \\
\bottomrule
\end{tabular}
\end{table}

We evaluate the state estimators on the physical Lima robot in two distinct scenarios, to gauge the sim-to-real gap and assess the real-time performance of the estimators with a policy. 

First, we evaluate them offline on a 10-minute ground-motion dataset comprising 20 messy motion sequences collected with a motion capture (MoCap) system. The MoCap pose data is fused with IMU measurements via a classical InEKF without any contact states to obtain the ground-truth position, orientation, and velocity estimates.

The linear velocity ATE in the root frame, as well as the position and orientation ATE in the world frame, are summarized in \tabref{tab:ate_total_comparison_ground_motion_real}. These real-world results agree with our simulation results in \tabref{tab:ate_total_comparison_ground_motion} and indicate a low sim-to-real gap of our approach. The lower error values compared to the simulated experiments can be explained by less extreme motions due to the physical setup, as well as potentially smaller IMU biases on the real system. Note, however, that this evaluation is performed offline on recorded real-world data; models marked with $^*$ cannot run in real time.

Second, we run the state estimators in the loop with a VMP~\cite{serifi_vmp_2024} controller for 13 dynamic dance motions. We utilize the same non-privileged policy inputs as in simulation. The policy inputs contain the body frame linear velocity estimates from our state estimator, alongside the body frame angular velocity directly extracted from the on-board IMU.
To demonstrate generalization, we include challenging unseen motions: a pirouette and a moonwalk. We also compare with the previously described MoCap-based state estimator that directly tracks the robot's IMU frame without contacts. Note that some baseline models were too computationally expensive for execution on the robot at the \SI{600}{\hertz} rate, and are therefore not evaluated here.

For each method, we attempt each of the seen motions $5\times$, and the unseen pirouette and moonwalk $10\times$. We record the success rate (i.e., when the robot does not fall). Results are summarized in \tabref{tab:success_chance_dancing}. It can be seen that our method outperforms all other methods, including the MoCap state estimation --- this highlights the challenging nature of the motions. We also refer to the supporting video.

\section{CONCLUSION}

CoCo-InEKF is a differentiable Invariant Extended Kalman Filter that utilizes a neural module to predict contact velocity covariances rather than relying on binary contact states. Trained end-to-end via backpropagation through time, the framework avoids the need for ground-truth contact labels and effectively handles complex contact states. Experiments on the Lima bipedal robot demonstrate that the method advances the accuracy-efficiency Pareto front, and can run within a \SI{600}{\hertz} onboard control loop. Furthermore, the system is insensitive to the exact placement of contact candidates, supporting an automated selection process that performs on par with expert-handpicked configurations.

While trained exclusively in simulation, the method supports training on real-world data with ground-truth states obtained from motion capture. We are eager to explore whether incorporating real-world data or greater training diversity can further improve performance. In the future, we plan to apply the framework to diverse robot morphologies and integrate these proprioceptive estimates with exteroceptive sensors, such as LiDAR or vision, for global drift correction. Such advancements will pave the way for agile robots capable of navigating unpredictable, contact-rich environments with unprecedented robustness.

%% Use plainnat to work nicely with natbib. 
\bibliographystyle{plainnat}
\bibliography{main}

% Appendix
% \appendix

\end{document}